\documentclass[review]{elsarticle}
\usepackage{hyperref}
\usepackage{float}
\usepackage{verbatim} 
\usepackage{subfig}
\usepackage{graphicx}
\usepackage{amsfonts}
\usepackage{algorithm2e}
\RestyleAlgo{ruled}
\usepackage{amsmath}
\restylefloat{figure}
\restylefloat{table}
\usepackage{bbm}
\usepackage{mathtools}
\usepackage{amssymb}
\usepackage{amsthm}
\usepackage{multirow}
\usepackage[table,xcdraw]{xcolor}

\DeclareMathOperator*{\argmax}{arg\,max}

\newtheorem{definition}{Definition}
\newtheorem{assumption}{Assumption}

\newtheorem{theorem}{Theorem}
\newtheorem*{proof*}{Proof}

\journal{Journal of Manufacturing Systems}

\begin{document}
\begin{frontmatter}

\begin{titlepage}
\begin{center}
\vspace*{1cm}

\textbf{ \large Self-optimization in distributed manufacturing systems using Modular State-based Stackelberg Games}

\vspace{1.5cm}

Steve Yuwono$^a$ (yuwono.steve@fh-swf.de), Ahmar Kamal Hussain$^b$ (ahmar.hussain@ovgu.de), Dorothea Schwung$^c$ (dorothea.schwung@hs-duesseldorf.de), Andreas Schwung$^a$ (schwung.andreas@fh-swf.de) \\

\hspace{10pt}

\begin{flushleft}
\small  
$^a$ Department of Automation Technology and Learning Systems, South Westphalia University of Applied Sciences, Lübecker Ring 2, Soest, 59494, Germany\\
$^b$ Data and Knowledge Engineering Group, Otto von Guericke University Magdeburg, Universit{\"a}tsplatz 2, Magdeburg, 39106, Germany\\
$^c$ Department of Artificial Intelligence and Data Science in Automation Technology, Hochschule D{\"u}sseldorf University of Applied Sciences, M{\"u}nsterstra{\ss}e 156, D{\"u}sseldorf, 40476, Germany

\vspace{1cm}
\textbf{Corresponding Author:} \\
Steve Yuwono \\
Department of Automation Technology and Learning Systems, South Westphalia University of Applied Sciences, Lübecker Ring 2, Soest, 59494, Germany \\
Tel: (49) 1779074949 \\
Email: yuwono.steve@fh-swf.de

\end{flushleft}        
\end{center}
\end{titlepage}

\title{Self-optimization in distributed manufacturing systems using Modular State-based Stackelberg Games}

\author[label1]{Steve Yuwono\corref{cor1}}
\ead{yuwono.steve@fh-swf.de}

\author[label3]{Ahmar Kamal Hussain}
\ead{ahmar.hussain@ovgu.de}

\author[label2]{Dorothea Schwung}
\ead{dorothea.schwung@hs-duesseldorf.de}

\author[label1]{Andreas Schwung}
\ead{schwung.andreas@fh-swf.de}

\cortext[cor1]{Corresponding author.}
\address[label1]{Department of Automation Technology and Learning Systems, South Westphalia University of Applied Sciences, Lübecker Ring 2, Soest, 59494, Germany}
\address[label3]{Data and Knowledge Engineering Group, Otto von Guericke University Magdeburg, Universit{\"a}tsplatz 2, Magdeburg, 39106, Germany}
\address[label2]{Department of Artificial Intelligence and Data Science in Automation Technology, Hochschule D{\"u}sseldorf University of Applied Sciences, M{\"u}nsterstra{\ss}e 156, D{\"u}sseldorf, 40476, Germany}

\begin{abstract}
In this study, we introduce Modular State-based Stackelberg Games (Mod-SbSG), a novel game structure developed for distributed self-learning in modular manufacturing systems. Mod-SbSG enhances cooperative decision-making among self-learning agents within production systems by integrating State-based Potential Games (SbPG) with Stackelberg games. This hierarchical structure assigns more important modules of the manufacturing system a first-mover advantage, while less important modules respond optimally to the leaders' decisions. This decision-making process differs from typical multi-agent learning algorithms in manufacturing systems, where decisions are made simultaneously. We provide convergence guarantees for the novel game structure and design learning algorithms to account for the hierarchical game structure. We further analyse the effects of single-leader/multiple-follower and multiple-leader/multiple-follower scenarios within a Mod-SbSG. To assess its effectiveness, we implement and test Mod-SbSG in an industrial control setting using two laboratory-scale testbeds featuring sequential and serial-parallel processes. The proposed approach delivers promising results compared to the vanilla SbPG, which reduces overflow by 97.1\%, and in some cases, prevents overflow entirely. Additionally, it decreases power consumption by 5-13\% while satisfying the production demand, which significantly improves potential (global objective) values.

\end{abstract}

\begin{keyword}
Stackelberg games, state-based potential games, game theory, reinforcement learning, modular production systems, production optimization
\end{keyword}

\end{frontmatter}

\section{Introduction}\label{sec:intro}

In modern industry, the integration of Industrial Cyber-Physical Systems (ICPS), automation, and artificial intelligence (AI) forms a transformative technology paradigm that improves manufacturing systems. ICPS combines computational and physical processes to optimize performance, while automation reduces human intervention, and AI enables machines to make autonomous decisions by learning from past experiences. Recent studies~\cite{Jan2023, Teerasoponpong2022} highlight the significant impact of these integrations, which results in improved efficiency, productivity, and adaptability. In the current industrial landscape, distributed manufacturing systems have gained popularity due to their decentralized nature, which offers greater flexibility and scalability~\cite{Mourtzis2020, ElMaraghy2021}. These systems, often modelled as multi-agent systems, involve numerous control variables with complex interrelations. Applications range from distributed control with industrial robots~\cite{Petr2022} to process optimization in material extrusion~\cite{Salmi2022}. Key advantages include improved flexibility, adaptability, fault detection, plug-and-play functionality, and responsiveness to dynamic production demands. However, challenges in coordination, optimization, and adaptability remain, which necessitate innovative solutions as system complexity increases.

The introduction of AI-driven self-optimization has revolutionized distributed manufacturing systems, which enables autonomous learning and real-time adaptability. These self-learning systems improve decision-making by considering past experiences or historical data, which results in reduced downtime, lower energy consumption, better resource utilization, and overall efficiency gains. Machine learning, particularly, plays a key role by identifying patterns in real-time data, which allows systems to adapt and respond proactively to changing conditions. Several studies demonstrate its effectiveness, including deep reinforcement learning (RL) for flexible job shop scheduling~\cite{Zhang2023}, adaptive PLC-based control for distributed production through model-based deep learning and model predictive control~\cite{Yuwono2023a}, and automatic PLC code generation using evolutionary algorithms~\cite{Loeppenberg2023}. However, real-world applications remain limited due to challenges such as computational constraints, unpredictable algorithmic behaviour, and solution stability.

Our recent investigations~\cite{Schwung2023, Yuwono2023b} indicate that self-learning, facilitated by a dynamic game theoretical (GT) approach~\cite{Marden2012, Zazo2016}, provides a robust solution for distributed learning. While GT has predominantly been applied to machine learning through the analysis of simultaneous play games~\cite{Fiez2020}, we have identified practical challenges in applying simultaneous decision-making for real-world control of multi-agent manufacturing systems. First, simultaneous learning lacks inherent coordination among agents, which can lead to conflicting decisions due to differing objectives. Second, in our previous work on State-based Potential Games (SbPG)~\cite{Schwung2022}, we found that equal-weighted learning is inadequate for production system control, where certain actuators, such as those that potentially create disruptions in production flow, are more critical than others. Additionally, theoretical considerations caution against exclusively relying on Nash equilibria, which highlights the broader applicability of Stackelberg equilibria in games characterized by convex costs and strategy spaces~\cite{Fiez2020, Basar1998}.

The challenges identified highlight the need for a hierarchical order of play, addressed in GT through Stackelberg games~\cite{Simaan1973, Bauso2016}. In this study, we integrate the effective concept of SbPG with Stackelberg games in modular systems, which results in a novel game structure. The Stackelberg equilibrium~\cite{Stackelberg2010} serves as the min-max solution in general-sum games. In cooperative Stackelberg games, players assume leader-follower roles to enhance interactions, facilitate cooperative decision-making, and optimize the global objective function. Thus, more critical actuators can be appointed as leaders, with others acting as followers. Leader as well as follower interactions are governed by SbPG, which has been proven to converge in~\cite{Schwung2022}. A Stackelberg game then determines the combined strategies of leaders and followers. Furthermore, we aim to ensure that the proposed game structure is compatible with self-optimizing algorithms, such as gradient-based learning~\cite{Yuwono2024a} and RL~\cite{Sutton2018}.

The main contributions of this paper are as follows:
\begin{itemize}
    \item We introduce a novel cooperative game structure for self-optimization in distributed manufacturing systems, termed Modular State-based Stackelberg games (Mod-SbSG), which combines hierarchical Stackelberg games with leader and follower structure with the concept of SbPG.
    \item We examine various configurations of Stackelberg games, which investigate scenarios with single and multiple leaders, as well as varying focuses for leaders and followers.
    \item We provide convergence guarantees for the proposed novel Mod-SbSG structure resulting in guidelines for the learning algorithms.
    \item We propose a novel learning concept for Mod-SbSG integrating cooperative learning within leader and follower groups with hierarchical learning of these subgroups.
    \item We validate proposed approaches in laboratory environments, specifically using the Bulk Good Laboratory Plant and its larger-scale counterpart, which shows significant improvements over the vanilla SbPG, with up to a 97.1\% reduction in bottlenecks and system overflow, along with a 5-13\% decrease in power consumption.
\end{itemize}

The paper is organized as follows: Sec.~\ref{sec:prem} reviews preliminary research relevant to our study. Sec.~\ref{sec:prob} outlines the problem descriptions. Sec.~\ref{sec:modsbsg} details the proposed Mod-SbSG concept, while Sec.~\ref{sec:convergence} provides the proof of convergence for the proposed method. Sec.~\ref{sec:algos} describes the learning algorithms and their dynamics. Sec.~\ref{sec:res} includes details of the testing environments and discusses the results, and we conclude the paper in Sec.~\ref{sec:conclusion}.

\section{Literature review}\label{sec:prem}
This section focuses on a discussion of preliminary research on self-learning manufacturing systems using AI and Stackelberg games for engineering applications.

\subsection{Self-learning manufacturing systems using artificial intelligence}\label{sec:prem_1}

Recent advancements in automation have rapidly transformed manufacturing systems, with AI playing a key role in improving operational efficiency through real-time decision-making. This synergy marks a new era in manufacturing, where AI-driven systems adapt swiftly to dynamically changing production demands and configurations that lead to a more responsive and agile industrial landscape. A significant contribution of AI is in enhancing quality control, which improves precision in inspection and early defect detection using computer vision~\cite{Yunjie2022} and machine learning~\cite{Kharitonov2022}. Beyond quality control, AI also optimizes production planning~\cite{Hameed2023, Zhaojun2023}, including scheduling, demand forecasting, resource allocation, and inventory management. Moreover, AI proves highly effective in robotics~\cite{Loeppenberg2024} and Human-Robot Collaboration~\cite{Hoffman2019}. The integration of AI in manufacturing not only reduces operational costs but also enhances safety and energy efficiency.

Furthermore,  AI-based systems have self-learning capabilities, which enable continuous optimization by adapting to changing data and conditions. This research focuses on distributed self-optimizing manufacturing systems. Prior studies have integrated machine learning into self-learning frameworks, including multi-agent RL~\cite{Sutton2018}, PLC-policy combinations with machine learning~\cite{Schwung2023, Schwung2021}, model predictive control enhanced by adaptive PLC-policy via model-based deep learning~\cite{Yuwono2023a}, and dynamic GT~\cite{Zazo2016}. Among these, the GT-based approach is the most applicable due to its robustness, proven convergence, and efficient computational time. Particularly, GT is well-suited for distributed multi-agent manufacturing systems, which involve multiple independently trained control variables. It facilitates suboptimal cooperation among agents to maximize global objectives instead of focusing solely on local objectives. One widely used method is the SbPG~\cite{Marden2012} with gradient-based learning algorithms~\cite{Yuwono2024a}, and its model-based variants~\cite{Yuwono2023b, Yuwono2023c}. However, both multi-agent RL and SbPG structures often demonstrate limited cooperation, as agents act independently without considering others’ actions. To address this limitation, we propose a novel game structure integrating Stackelberg games to enhance decision-making in self-learning algorithms.

\subsection{Stackelberg games for engineering applications}\label{sec:prem_2}

Stackelberg games~\cite{Simaan1973} involve a sequential decision-making process, which contrasts with simultaneous games. In these games, the leader makes the initial move, followed by followers who react based on the leader's actions. While typically associated with non-cooperative GT, known as Stackelberg competition~\cite{Etro2008}, these games can also be adapted for cooperative scenarios. This interaction aims to achieve global objectives or Stackelberg equilibria~\cite{Stackelberg2010}. As GT's application in engineering expands, the use of Stackelberg games has similarly increased. Several examples include their deployment in anti-jamming defence for wireless networks~\cite{Luliang2018}, power control communications~\cite{Yan2014}, addressing security issues in networked control systems as defender-attacker games~\cite{Yuzhe2018}, and developing Stackelberg Actor-Critic methods in RL~\cite{Liyuan2022}.

Despite the wide-ranging applications of Stackelberg games in engineering, their utilization in self-learning manufacturing systems remains limited. Therefore, this research aims to address this gap by integrating Stackelberg games into the self-learning domain to enhance collaboration among players. The sequential decision-making characteristic of Stackelberg games sets them apart from other game types, such as dynamic potential games~\cite{Zazo2016} or multi-agent RL~\cite{Sutton2018}, where agents select actions simultaneously. This distinctive feature becomes a focal point in exploring the benefits of Stackelberg games in this domain. Currently, no strategic game structure, apart from SbPG~\cite{Schwung2022}, supports distributed self-learning algorithms. However, SbPG also operates on simultaneous actions, which can result in unequal treatment of critical players. To address this, we propose a novel game structure, Mod-SbSG, which combines Stackelberg games with SbPG, which contains three distinct games, i.e. SbPG among leaders, SbPG among followers, and Stackelberg games between both coalition strategies. Additionally, we investigate the integration of this structure with gradient-based learning~\cite{Yuwono2024a} and the Advantage Actor-Critic (A2C) algorithm~\cite{Mnih2016}, which aims to reveal the potential benefits of cooperative leader-follower games in distributed multi-agent systems.

\section{Problem description}\label{sec:prob}

\begin{figure}[ht]%
    \centering
    \includegraphics[width=1.00\linewidth]{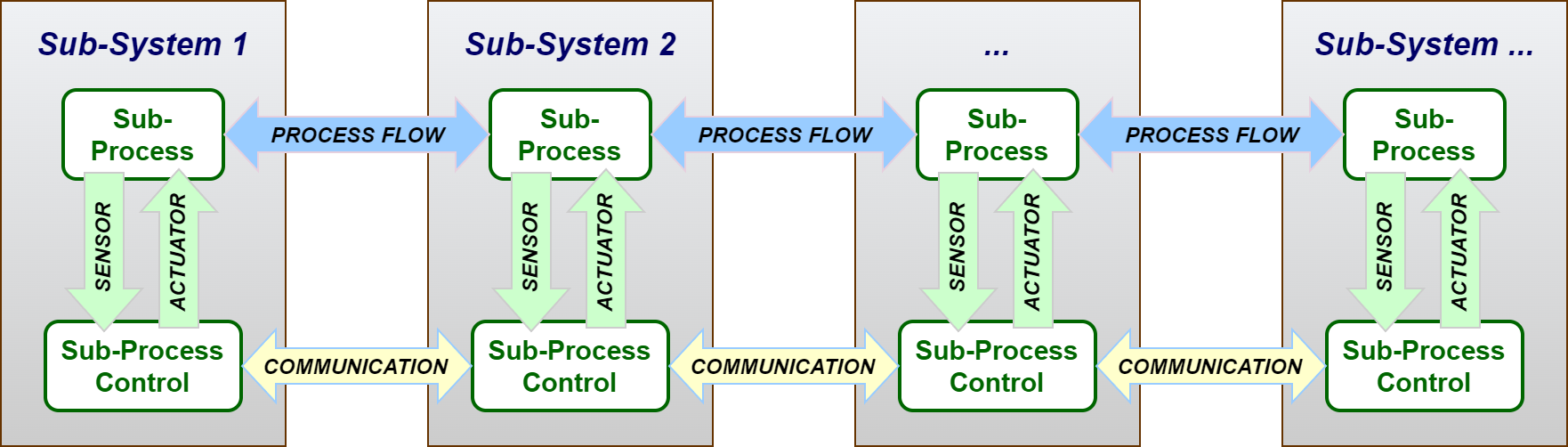}
    \caption{An illustration of modular production units~\cite{Yuwono2024b}.}%
    \label{fig:mod_systems}%
\end{figure}

This section outlines the problem addressed in this study, which focuses on developing autonomous optimization methods for fully distributed manufacturing systems. Specifically, we focus on modular systems comprising multiple subsystems, each with its own local control system and potentially distinct objectives, as depicted in Fig.~\ref{fig:mod_systems}. These subsystems are interconnected through either parallel or serial-parallel configurations and interact with their control systems via the exchange of local signals. Each subsystem contains one or more actuators, which can be conceptualized as individual players $i$ in GT terms. Our main goal is to facilitate self-optimization across these systems in a distributed manner, thereby eliminating the requirement for centralized control and allowing for flexible, scalable, and reusable operations across diverse modules through instantiation.

\begin{figure}[ht]%
    \centering
    \includegraphics[width=1.00\linewidth]{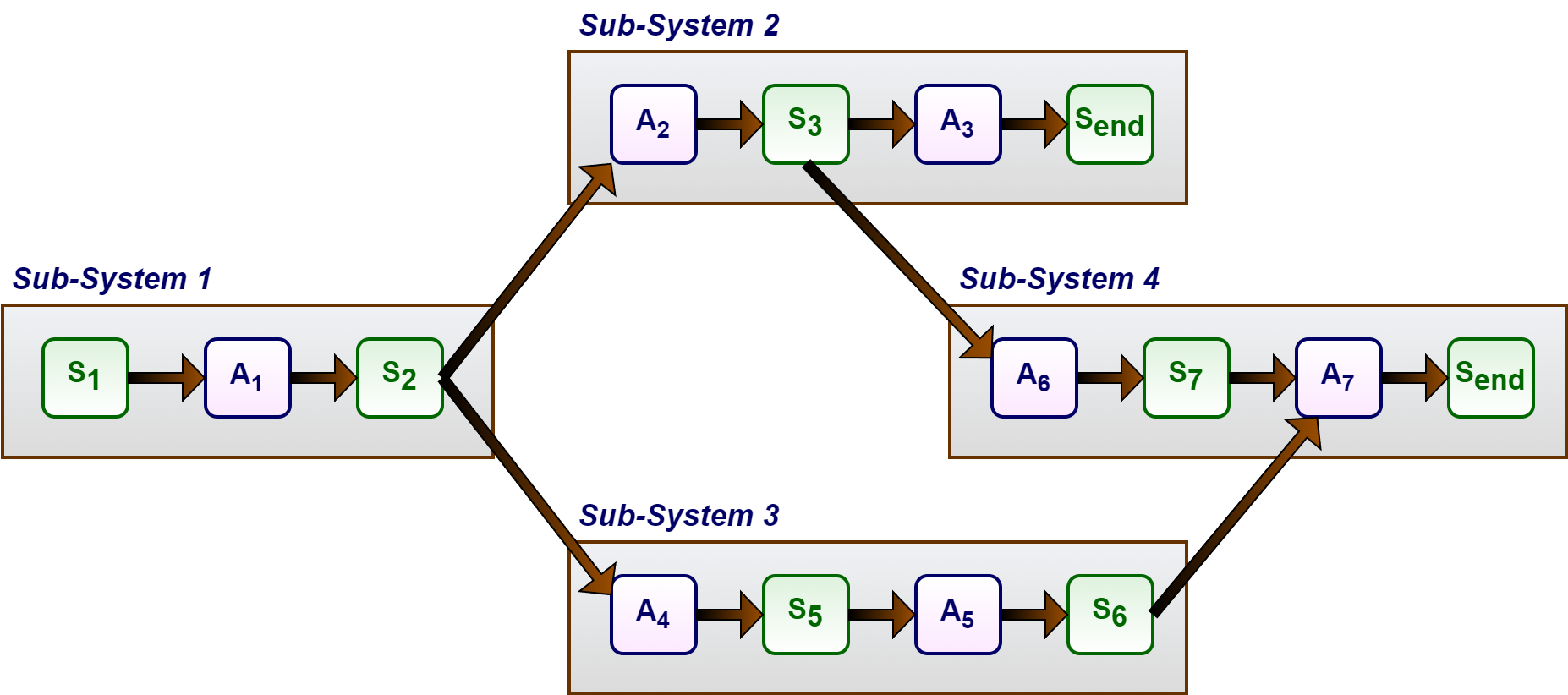}
    \caption{A schematic diagram of a production chain featuring serial-parallel connected subsystems.}%
    \label{fig:chain_systems}%
\end{figure}

We follow graph theory~\cite{Yamamoto2015} to model the distributed system, which constitutes a production chain as explained in~\cite{Schwung2022}. This production chain is represented in both serial and serial-parallel configurations, which features an alternating sequence of actuators (e.g., rotary feeders, motors, pumps, conveyors, and more) and physical states that indicate the process status, as illustrated in Fig.~\ref{fig:chain_systems} for serial-parallel processes. These actuators are anticipated to display a hybrid actuation system with both continuous and discrete operational behaviours.

The production chain is modelled as a dynamic sequence involving actuators $\mathcal{N}={1,\ldots,N}$, with each actuator associated with sets of continuous or discrete actions ${A}_i \subset \mathbb{R}^c \times \mathbb{N}^d$, and a set of states $\mathcal{S}\subset \mathbb{R}^m$. In this model, the edges $\mathcal{E}$ of the graph do not include connections of the form $e=(A_i,A_j)$ and $e=(s_i,s_j)$, where $A_i, A_j \in \mathcal{N}$ and $s_i, s_j \in \mathcal{S}$. For each actuator $A_i\in \mathcal{N}$, we define two sets of neighbouring states, which are the preceding neighbour states $\mathcal{S}^{A_i}_{prior}=\{s_j \in \mathcal{S}|\exists e=(s_j,A_i)\in \mathcal{E} \}$ and the subsequent neighbour states $\mathcal{S}^{A_i}_{next}=\{s_j \in \mathcal{S}|\exists e=(A_i,s_j)\in \mathcal{E}\}$.

To be noted, assuming the production chain consists only of sequences of states and actions is not overly restricting. More complex arrangements involving multiple states can be represented as a unified state vector, and the same applies to actions. This distributed production model is applicable across numerous sectors in the process industry, such as food production, oil and gas, chemical manufacturing, pharmaceuticals, and water treatment.

Since we are addressing optimization problems in real manufacturing systems, each subsystem may have specific objectives. In GT terms, these objectives can be represented as utilities. Therefore, we assume that each player $i$ has a local utility function  $U_i(a_i, S^{A_i})$, where ${S}^{A_i} \in \mathcal{S}^{A_i} = \mathcal{S}^{A_i}_{prior} \cup \mathcal{S}^{A_i}_{next} \cup \mathcal{S}^{g}$, with $\mathcal{S}^{g}$ expressing the states associated with global objectives. 

Consequently, we aim to maximize the overall system utility $\phi$
\begin{equation} \label{Eq:overall_util}
\max_{a_i\in {A}_i} \phi({a},{S}),
\end{equation}
by jointly maximizing local utilities $U_i(a_i, S^{A_i})$. Note that the above optimization problem is formulated in a fully distributed manner in the sense, that optimizing local utilities results in the optimization of the overall utility.

A significant challenge arising from the problem description is managing the priority among players. Each player $i$ has specific objectives represented by a local utility function $U_i(a_i, S^{A_i})$. However, the impact of each local utility function $U_i$ on the global objective $\phi$ cannot be considered equal, as some players have a greater influence than others. Additionally, in the context of a production chain, the actions of each player $a_i$ affect the states of their surroundings, which in turn influences the utility values of surrounding players. In our previous research~\cite{Yuwono2023b, Schwung2022, Yuwono2024a, Schwung2021, Yuwono2024c}, we did not address player prioritization. Instead, we treated all players equally and allowed them to engage in simultaneous games. In this study, we aim to address the issue of player prioritization by employing leader-follower games, which leads to a novel game structure. Leaders will be able to play simultaneous games among themselves, as will the followers, while interactions between leaders and followers will be defined according to Stackelberg's strategies. This novel game structure results in a change of the underlying distributed optimization problem resulting in considerable improvement of results.

\section{Modular State-based Stackelberg Games}\label{sec:modsbsg}

In modular production units, players that significantly impact the outputs of their surroundings and the global objective (potential) function are considered more critical. Typically, these critical players are positioned higher in the hierarchy, often serving as leaders. Consequently, treating all players equally as is current state of the art, may not be the most effective approach. Contrary, we propose to assign each player a role as either a leader or a follower, with leaders being the more critical players. This results in a group of leader and a group of follower modules.

Consequently, we propose the Mod-SbSG as a game structure composed of three key sub-games: (1) a cooperative game for the group of leader modules, (2) a cooperative game for the group of follower modules, and (3) a hierarchical game governing the interactions between the leader and follower groups.
\begin{figure}[ht]%
    \centering
    \includegraphics[width=1.00\linewidth]{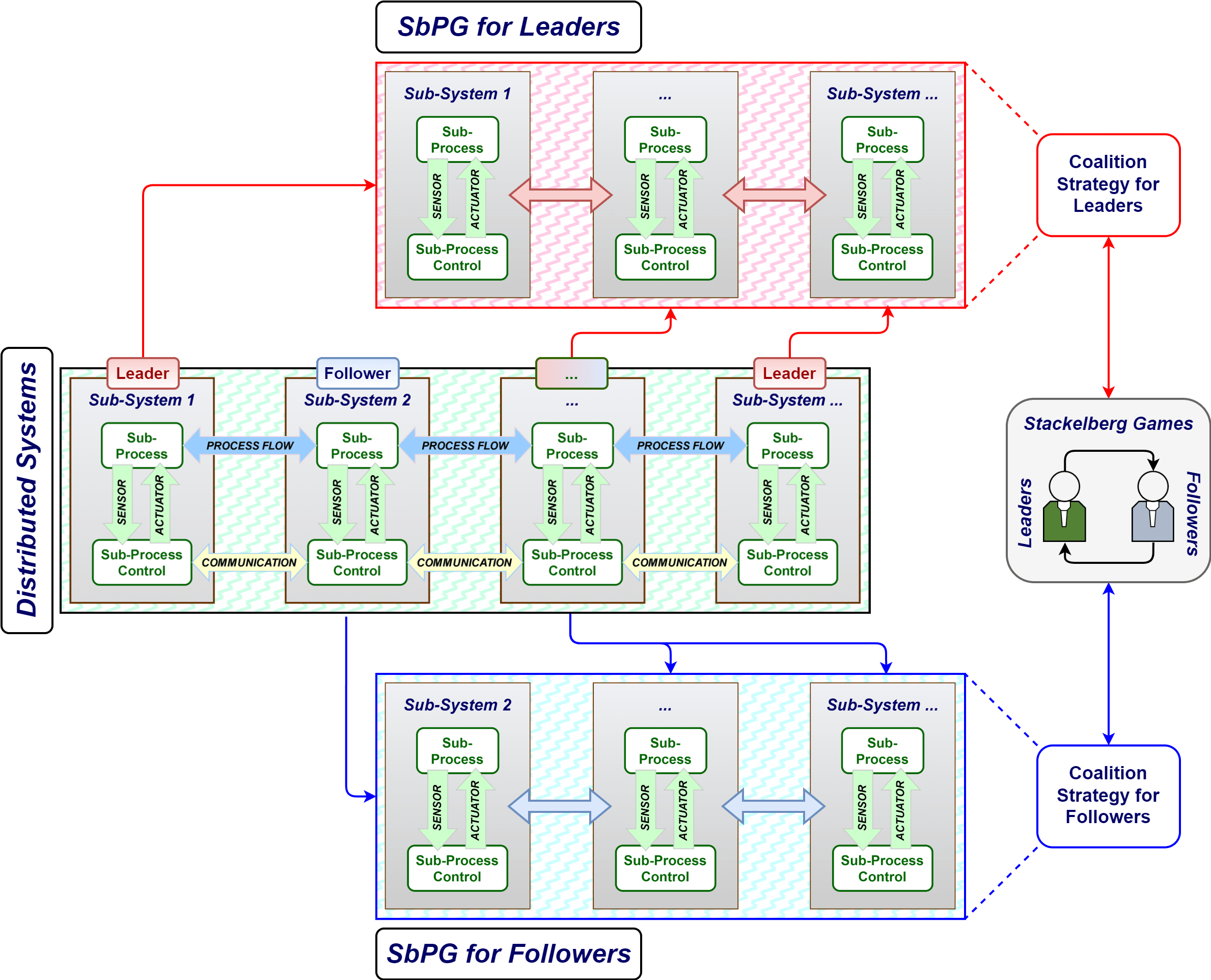}
    \caption{An overview of Mod-SbSG in distributed manufacturing systems.}%
    \label{fig:overview}%
\end{figure}

Specifically, for the cooperative games of leader and follower groups, we propose to set up an SbPG among each group, which has been proven effective and convergent for distributed systems in~\cite{Schwung2022}. The hierarchical game describing the interactions between the leader and follower groups is addressed using a Stackelberg game. This game structure is detailed in the following subsections.

\subsection{SbPG for leader and follower groups}\label{sec:modsbsg_1}

We propose to use SbPG for the coordination game within the group of leaders and followers respectively. Potential games~\cite{Monderer1996} provide a game structure for modelling and studying strategic interactions between rational players, where each player's utility $U_i$ is influenced by both their actions and the state of the environment. These interactions are then assessed by a scalar potential function $\phi$, which acts as a global objective. SbPG~\cite{Marden2012} extend this framework by explicitly including state information in the players' strategic decision-making process.

SbPGs are further extended in~\cite{Schwung2022} to manage self-optimizing modular production units by incorporating the set of states $S$ and the state transition process $P$. The formal definition of an SbPG for the group of leaders following the approach of~\cite{Schwung2022} with $l\in\mathcal{L}=\{1,2,\ldots,L\}$, where $L$ denotes the total number of leaders, is presented as follows:
\begin{definition}\label{def:sbpg_leader}
    A game $\Gamma_L(\mathcal{L}, \mathcal{A}_L, \{U_l\}, {S}, P, \phi_L)$ is an SbPG for leaders if it meets the following conditions for the potential function:
    \begin{align}\label{eq:potcondsbpg_leader}
    U_l(a_l,s) - U_l({a}^{\prime}_l, {a}_{-l},s) = \phi_L(a_l,s) - \phi_L({a}^{\prime}_l, {a}_{-l},s),
    \end{align}
    and
    \begin{align}\label{eq:condsbpg_leader}
    \phi_L(a_l,s^{\prime}) \geq \phi_L(a_l,s),
    \end{align}
    for any state $s^{\prime}$ in $P(a,s)$.
\end{definition}

Similarly, the SbPG for followers with $f\in\mathcal{F}=\{1,2,\ldots,F\}$, where $F$ denotes the total number of followers, is defined as follows:
\begin{definition}\label{def:sbpg_follower}
    A game $\Gamma_F(\mathcal{F}, \mathcal{A}_F, \{U_f\}, {S}, P, \phi_F)$ is an SbPG for followers if it satisfies the following conditions for the potential function:
    \begin{align}\label{eq:potcondsbpg_follower}
    U_f(a_f,s) - U_f({a}^{\prime}_f, {a}_{-f},s) = \phi_F(a_f,s) - \phi_F({a}^{\prime}_f, {a}_{-f},s),
    \end{align}
    and
    \begin{align}\label{eq:condsbpg_follower}
    \phi_F(a_f,s^{\prime}) \geq \phi_F(a_f,s),
    \end{align}
    for any state $s^{\prime}$ in $P(a,s)$.
\end{definition}
Note, that the above SbPG for leaders and followers operates as a closed game in the sense, that no interactions take place between the individual game structures. Hence, with a suitable learning algorithm, both games independently converge to their corresponding Nash equilibria according to the convergence guarantees discussed in Sec.~\ref{sec:convergence}.

\subsection{Leader-follower game as a Stackelberg game}\label{sec:modsbsg_3}

Stackelberg games~\cite{Simaan1973} model strategic interactions where one or more players act as leaders, and the others as followers. Unlike simultaneous-move games, Stackelberg games feature a sequential decision-making process. The leader makes the first move, followed by the followers who react sequentially based on the leader's actions. This hierarchical structure gives the leader a strategic advantage, which allows them to optimize their decisions and maximize their objectives by anticipating the predictable responses (best responses) of the followers. Subsequently, the follower recognises the leader's decisions and formulates responses based on their own strategic considerations. Although often applied in non-cooperative games, hierarchical decision-making in Stackelberg games can be adapted to cooperative frameworks. Techniques like dynamic programming, variational inequalities, and Stackelberg equilibrium concepts~\cite{Stackelberg2010} are commonly used to analyze and solve for equilibrium outcomes in these settings.

Hence, after appointing the players as leaders and followers and managing the SbPG for each group, we propose to manage the interaction between the leaders' coalition strategy, $\mathcal{A}_L$, and the followers' coalition strategy, $\mathcal{A}_F$ by means of a Stackelberg game~\cite{Fiez2020} as outlined below:
\begin{definition}\label{def:sg_1_1}
    Consider a game $\Gamma_S(\mathcal{N}, \mathcal{A}, \phi_L, \phi_F)$ with a set of players $\mathcal{N}:\mathcal{L}\times\mathcal{F}$ consisting of a leader group $\mathcal{L}$ and a follower group $\mathcal{F}$ and their combined action space $\mathcal{A}=\mathcal{A}_L\times \mathcal{A}_F\in\mathbb{R}^{m}$, where $\mathcal{A}_L = a_1 \times a_2 \times \ldots \times a_{m_L} \in\mathbb{R}^{m_L}$ and $\mathcal{A}_F = a_1 \times a_2 \times \ldots \times a_{m_F} \in\mathbb{R}^{m_F}$. Further, we define an objective function $\phi_L:\mathcal{A}_L\rightarrow\mathbb{R}$ for the leader group and an objective function $\phi_F:\mathcal{A}_F\rightarrow\mathbb{R}$ for the follower group. This game is called a cooperative Stackelberg game, if the following optimization problem is solved:
    \begin{equation}\label{eq:sg_1_1}
        \max_{{a_L}\in \mathcal{A}_L} \{\phi_L({a_L},{a_F}) | a_F \in \argmax_{{y}\in \mathcal{A}_F}\phi_F({a_L},{y})\},
    \end{equation}
    \begin{equation}\label{eq:sg_1_2}
        \max_{{a_F}\in \mathcal{A}_F} \phi_F({a_L}, {a_F}).
    \end{equation}
\end{definition}
Note that we can use the potential functions $\phi_L$ and $\phi_F$ as objective functions within the Stackelberg game due to the specific properties of the SbPG in Eq.~\eqref{eq:potcondsbpg_leader} and~\eqref{eq:potcondsbpg_follower}. 
Also note that the Stackelberg game is defined for the coalition strategies of the two roles, regardless of the number of leaders or followers. This means that whether there is a single leader with multiple followers or multiple leaders and followers, the Stackelberg game effectively involves only two players: one representing the leaders' strategy $a_L$ and the other representing the followers' strategy $a_F$. Following the two-player Stackelberg game formulation from~\cite{Fiez2020}, we discuss convergence properties in Sec.~\ref{sec:convergence}.

\subsection{Overall game structure}\label{sec:modsbsg_4}

After defining the individual games of leader and follower as well as the Stackelberg game to connect these two groups, we formulate the game structure of Mod-SbSG as below:
\begin{definition}\label{def:Mod-SbSG}
    A game $\Gamma(\mathcal{N, L, F}, {A}, \{u_i\}, \mathcal{S}, \mathcal{P}, \{\phi_L, \phi_F\})$ is called a Mod-SbSG, if the decision-making within leaders and followers is governed by the two SbPGs $\Gamma_L(\mathcal{L}, \mathcal{A}_L, \{U_l\}, {S}, P, \phi_L)$ and $\Gamma_F(\mathcal{F}, \mathcal{A}_F, \{U_f\}, {S}, P, \phi_F)$, while interactions between leaders and followers are modelled as Stackelberg game $\Gamma_S(\mathcal{N}, \mathcal{A}, \phi_L, \phi_F)$.
\end{definition}

\section{Convergence analysis}\label{sec:convergence}

After the definition of Mod-SBSG, we now focus on the convergence properties of the overall game structure. To this end, we have to consider the different game structures, namely SbPG as well as Stackelberg games. More specifically, we first analyse the convergence properties of the SbPG, and subsequently, the convergence properties of the Stackelberg game under the assumption, that the SbPG converged to their respective equilibria.

For SbPG, there already exists a line of results with respect to their convergence properties. Specifically, it has been shown that exact potential games converge to a Nash equilibrium under best-response dynamics~\cite{Marden2012}. Under some mild conditions, this result can be expanded to SbPG. Particularly, Zazo et al.~\cite{Zazo2016} establish criteria for proving the existence of an SbPG and demonstrate that it converges as long as these conditions are fulfilled. Based on these results, Schwung et al.\cite{Schwung2022} provide assumptions on the design of the utility functions, such that the distributed optimization of modular production units can be cast as an SbPG from which convergence guarantees follow directly. 

As we employ the exact same utility functions as in~\cite{Schwung2022} fulfilling their Assumptions 1-4, we can state the following theorem for the leaders' coalition game:
\begin{theorem}\label{theorem_1}
    Given Assumptions 1-4 from~\cite{Schwung2022}, the cooperative game between leaders, $\Gamma_L(\mathcal{L}, \mathcal{A}_L, \{U_l\}, {S}, P, \phi_L)$, as defined in Def.~\ref{def:sbpg_leader} constitute an SbPG.
\end{theorem}

\begin{proof*}\label{proof_1}
    The proof follows directly from Proposition 1  in~\cite{Schwung2022}.\hfill \qed
\end{proof*}

Similar to the leaders' game, we can state the following theorem for the followers' coalition game:
\begin{theorem}\label{theorem_2}
    Given Assumptions 1-4 from~\cite{Schwung2022}, the cooperative game between followers, $\Gamma_F(\mathcal{F}, \mathcal{A}_F, \{U_f\}, {S}, P, \phi_F)$, as defined in Def.~\ref{def:sbpg_follower} constitute an SbPG.
\end{theorem}
\begin{proof*}\label{proof_2}
    The proof follows directly from Proposition 1 in~\cite{Schwung2022}.\hfill \qed
\end{proof*}
Theorem~\ref{theorem_1} and~\ref{theorem_2} state that since modular production systems, in general, can be cast as SbPGs, so can subgames consisting of just leader modules and follower modules. The existence of the SbPG then comes with convergence guarantees resulting in convergence to a Nash equilibrium of the leader as well as the follower groups. 

Backed with the above results, we now focus on analyzing the convergence properties of the Stackelberg game. To this end, we first recall the definition of the differential Stackelberg equilibrium from~\cite{Fiez2020}:
\begin{definition}\label{def:Diff_Stackelberg}
    The pair $(a_L^*,a_F^*)\in \mathcal{A}$ with $a_F^*=r(a_L^*)$, where $r$ is implicitly defined by $\frac{\partial \phi_F(a_L^*,a_F^*)}{\partial a_F}=0$, is a differential Stackelberg equilibrium for the game $(\phi_1,\phi_2)$ with player 1 as the leader, if $\frac{d \phi_F(a_L^*,r(a_L^*))}{d a_L}=0$, and $\frac{d^2 \phi_L(a_L^*,r(a_L^*))}{d a^2_F}$ is positive definite.
\end{definition}
Note that the differences between the conditions for differential Stackelberg equilibria and the corresponding differential Nash equilibria described by the conditions $\left(\frac{\partial \phi_L(a^*)}{\partial a_L},\frac{\partial \phi_F(a^*)}{\partial a_F}\right)=0$ and $\frac{\partial^2 \phi_i(a^*)}{\partial a^2_i}>0$ for $i=L,F$, which is due to the implicitly defined best response $r(a_L^*)$ of the follower resulting in the use of the total derivative in Definition~\ref{def:Diff_Stackelberg}.   

Furthermore, we introduce a gradient-based update law of leader and follower as in~\cite{Fiez2020} which is defined as follows:
\begin{align}\label{eq:learningupdate1}
a_{k+1} = a_{k} - \alpha_{k} \left(\left(\frac{d \phi_L(x)}{d a_L},\frac{\partial \phi_F(x)}{\partial a_F}\right)^T\right)+\gamma_{k}, \quad k=0,1,\ldots
\end{align}
where $a_{k}=(a_{L,k},a_{F,k})^T$, $\alpha_{k}$ and $\gamma_{k}$ denote the sequence of learning rates and the exploration noise process, respectively, and
\begin{align}\label{eq:learningupdate2}
\frac{d \phi_L(x)}{d a_L} = \frac{\partial \phi_L(x)}{\partial a_L} - \frac{\partial \phi_L(x)}{\partial a_F} \left(\frac{\partial^2 \phi_F(x)}{\partial x^2_F}\right)^{-1} \frac{\partial^2 \phi_F(x)}{\partial a_L\partial a_F}.
\end{align}
To provide the convergence analysis, we have to make the following assumptions:
\begin{assumption}\label{ass:gb_sg}
    We assume that the gradient-based update of Eq.~\eqref{eq:learningupdate1} - \eqref{eq:learningupdate2} is used to update the Stackelberg game of Mod-SbSG.
\end{assumption}
\begin{assumption}~\cite{Fiez2020}\label{ass:conv}
    The following conditions hold:
    \begin{enumerate}
        \item The maps $\frac{d \phi_L}{d a_L} : \mathbb{R}^d \rightarrow \mathbb{R}^{d_1}$, $\frac{\partial \phi_F}{\partial a_F} : \mathbb{R}^d \rightarrow \mathbb{R}^{d_2}$ are $L_1$, $L_2$-Lipschitz, and $||\frac{d \phi_L}{d a_L}|| \leq M_1 < \infty$.
        \item For each $i\in \mathcal{N}$, the learning rates must satisfy the conditions $\Sigma_k \alpha_{i,k} = \infty$, $\Sigma_k \alpha_{i,k}^2 < \infty$.
        \item The noise processes $\{\gamma_{i,k}\}$ are zero mean martingale difference sequences. Specifically, given the filtration $\mathcal{F}_k = \sigma(a_s, \gamma_{1,s}, \gamma_{2,s}, s \leq k)$, ${\{\gamma_{i,k}\}}_{i\in \mathcal{N}}$ are conditionally independent, $\mathbb{E}[\gamma_{i,k+1}|\mathcal{F}_k] = 0$ a.s., and $\mathbb{E}[||\gamma_{i,k+1}||\, |\mathcal{F}_k]\, \leq c_i(1+||a_k||)$ a.s. for some constants $c_i \leq 0$, $i\in \mathcal{N}$.
    \end{enumerate}
\end{assumption}
\begin{assumption}~\cite{Fiez2020}\label{ass:conv2}
    For every $a_L$, $\dot{a}_F = - \frac{\partial \phi_F(a_L,a_F)}{\partial a_F}$ has a globally asymptotically stable equilibrium $r(a_L)$ uniformly in $a_L$ and $r:\mathbb{R}^{d_1} \rightarrow \mathbb{R}^{d_2}$ is $L_r$-Lipschitz.
\end{assumption}
Assumption~\ref{ass:conv} basically follows from the typical technical requirements of statistical learning theory and is easy to fulfil. Assumption~\ref{ass:conv2} is somewhat restrictive as it requires the follower SbPG, which defines the dynamics of $\dot{x}_2$ in the case of the Mod-SbSG, to have a global asymptotically stable Nash equilibrium. Particularly, SbPG typically exhibits multiple local Nash equilibria. However, extending our convergence results to local convergence is straightforward~\cite{Fiez2020} and omitted for brevity. 

We can now present the following theorem for validating the interaction between leaders and followers within the Stackelberg game which is an adjusted version of Proposition 8 in~\cite{Fiez2020}:
\begin{theorem}\label{theorem_3}
    Suppose that for each $a \in \mathcal{A}$ of the Mod-SbSG, $\frac{\partial^2 \phi_F}{\partial a^2_F}$ is non-degenerate, Assumptions~\ref{ass:gb_sg} and~\ref{ass:conv} hold and Assumption~\ref{ass:conv2} holds for the leader $i=L$. Then, $a_{L,k}$ converges almost surely to an equilibrium point $a_L^*$ which is a local Stackelberg solution for the leader. Moreover, if Assumption~\ref{ass:conv} holds for the follower $i=F$ and Assumption~\ref{ass:conv2} holds, then $a_{F,k} \rightarrow x_2^*=r(x_1^*)$ so that $(x_1^*,x_2^*)$ is a differential Stackelberg equilibrium.
\end{theorem}
\begin{proof*}\label{proof_3}
    The proof mainly follows the proof of Proposition 8 in~\cite{Fiez2020}. Particularly, it largely follows from known stochastic approximation results as Eq.~\eqref{eq:learningupdate1} - \eqref{eq:learningupdate2} are stochastic approximations of $\dot{a}_L = - \frac{d \phi_L(a_L,a_F)}{d a_L}$ which track the ODE asymptotically. Furthermore, as we employ the SbPG to define the dynamics of the follower, we can ensure convergence of the followers' dynamics with a non-degenerate $\frac{\partial^2 \phi_F}{\partial a^2_F}$. \hfill \qed
\end{proof*}
Note that the convergence behaviour is dependent on the employed training algorithms. Particularly, the training of the Stackelberg game has to be conducted by using the gradient-based update provided in Assumption~\ref{ass:gb_sg}. For the training of the subordinate SbPGs, we can operate either best-response learning~\cite{Schwung2022} or gradient-based learning~\cite{Yuwono2024a} as both have been proven to converge to local equilibria. A detailed explanation of the learning dynamics will be provided in the next section.

\section{Learning dynamics}\label{sec:algos}

After developing the game structure and discussing its convergence, we have to derive suitable learning algorithms for training the policies of each player, $\pi_l, \pi_f \in \pi_i$. To this end, we consider the learning dynamics induced by both the SbPG and the Stackelberg game. 

In~\cite{Schwung2022}, we initially proposed best-response learning for the SbPG structure, which utilizes ad-hoc random uniform sampling during the learning process. However, this random sampling approach led to lower predictability and potential instability in the learning process, as it lacked control over the learning direction. To address this issue, we improved the method by proposing gradient-based learning in~\cite{Yuwono2024a}, which provides guided learning and enables more stable convergence toward global optima compared to random sampling.

Both of these methods were originally designed for simultaneous games. However, Mod-SbSG introduces a hierarchical structure of leaders and followers within the player set $\mathcal{N}$, which necessitates a more complex dynamic. Given the more complex game structure, we have to address three key steps:
\begin{enumerate}
    \item the learning algorithm used for both leader and follower SbPG,
    \item the Stackelberg updates between the coalition strategies of the leaders and followers, and
    \item coordination of the learning dynamics, where we have to particularly address the update sequences as the Stackelberg game requires the follower group to converge before updating the leader group.
\end{enumerate}
As the Stackelberg game requires a gradient-based update for convergence, we propose to use a gradient-based update for all game structures. The policies for both learning algorithms in this study are represented in the form of performance maps as proposed in~\cite{Schwung2022}.

In what follows, we will define a formal representation of the players' policies, outline the learning update rule for leaders and followers, derive an approximate gradient descent algorithm to accommodate the data-driven nature of the game, develop a method for multi-step optimization for followers, and present the complete learning mechanism of Mod-SbSG.

\subsection{Policy representation using performance maps}\label{sec:algos_0}

\begin{figure}[t]
 \centering
 \includegraphics[width=0.80\columnwidth,keepaspectratio]{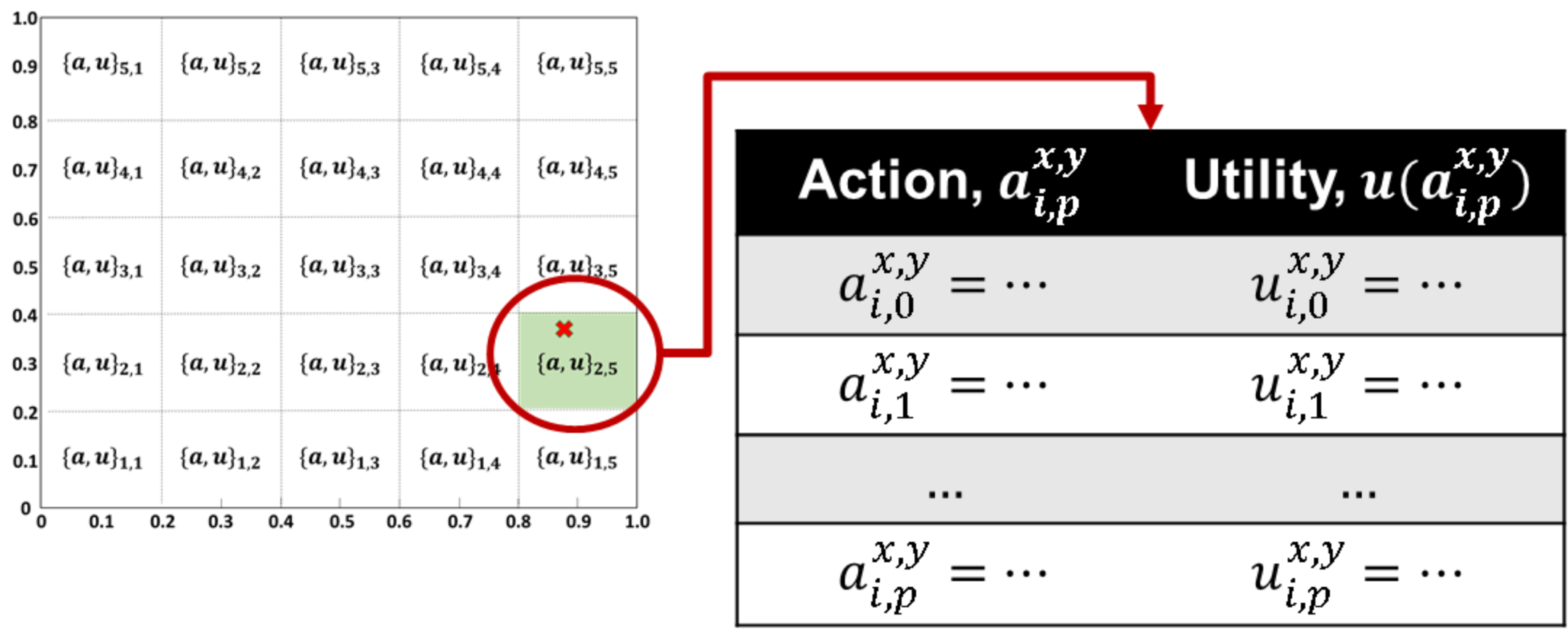}
\caption{5 x 5 performance map representation within a 2D state space in SbPG~\cite{Yuwono2024a}.}
\label{fig:per_maps}
\end{figure}

We begin by considering the representation of each player’s policy $\pi_l, \pi_f \in \pi_i$, which is responsible for storing the learned knowledge over various state-action pairs. In SbPG, the state space is discretized into equidistant support vectors, denoted by $q=1,\ldots,p$, which store the best-explored actions and their corresponding utility values for each state combination. Additionally, a stack of selected actions and their utilities is stored within each data point across different state combinations, as suggested in~\cite{Yuwono2024a}. Fig.~\ref{fig:per_maps} displays the performance map for each player $i$ in a system characterized by two states, $x$ and $y$.

Each player $i$ determines the next action $a_{i,t+1}$ by globally interpolating their performance map based on the current state $s_{i,t}$~\cite{Schwung2022}. The global interpolation process is as follows:
\begin{equation} 
    w_i^{\overline{s^0s^q}} = \dfrac{1}{{(d_i^{s^0s^q})}^2 + \gamma_{map}},
\end{equation}
\begin{equation} 
    a_{i} = \sum_{q} \dfrac{w_i^{\overline{s^0s^q}}}{\sum_{q}w_i^{s^0s^q}} \cdot a_i^q,
\end{equation}
where $s^0$ is the current state, $s^q$ refers to the state of the $q$-th support vector, $d_i^{s^0s^q}$ represents the absolute distance between $s^0$ and $s^q$, $w_i^{\overline{s^0s^m}}$ is the computed weight, and $\gamma_{map}$ is a smoothing parameter.

In Mod-SbSG, performance maps are required for both leaders and followers. For the leaders, the performance maps from the original approach in~\cite{Schwung2022} remain unchanged. This is because the input to each leader's policy $\pi_l$, used to compute its action $a_l$, relies solely on the state information $s_l$, which is equivalent to the input for each player $i$ in the vanilla SbPG structure.

A key distinction is in the performance maps for the followers, where each follower's action $a_f$ is determined not only by the state information $s_f$ but also by the coalition actions of the leaders $\mathcal{A}_L$. To deal with this additional input, we propose two potential solutions, either augmenting the performance map with extra dimensions or utilizing a stacking method. Upon evaluation, the first approach may limit the players' ability to fully explore the entire state space. Even if complete exploration is technically achievable, it would be time-intensive, which potentially leaves some grid cells unexplored and ultimately reduces the accuracy of the interpolation calculations.

\begin{figure}[ht]%
    \centering
    \includegraphics[width=0.75\linewidth]{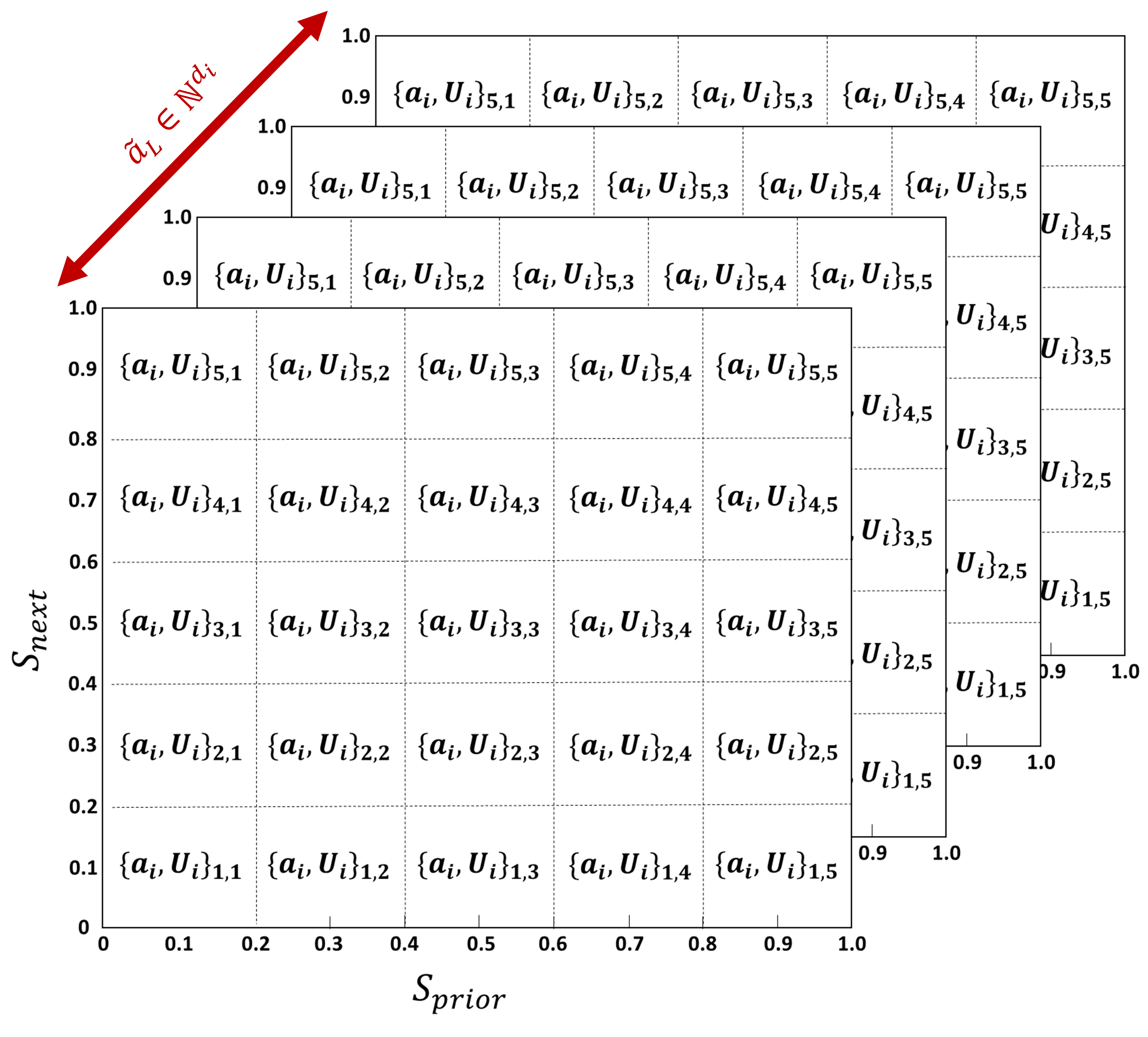}
    \caption{The updated structure of performance maps for the followers in Mod-SbSG on SbPG.}%
    \label{fig:maps_update}%
\end{figure}

Thus, the stacked performance map approach is favoured, as shown in Fig.~\ref{fig:maps_update}. This preference is due to its lower exploration requirements. When a follower interpolates its map, it references a specific layer corresponding to the selected leader's actions, rather than interpolating through a much larger, high-dimensional map. This method simplifies the process and reduces computational complexity.

\subsection{Learning update rule}\label{sec:algos_1}

Once the policy representation using performance maps is established, the next step is to update the action values in the support vectors and train the policies by designing a suitable training law. As previously mentioned, in the first update step of Mod-SbSG, both followers and leaders individually play SbPG using the gradient-based learning approach proposed in~\cite{Yuwono2024a}. Following this, the second step involves defining the Stackelberg rule within Mod-SbSG for allowing hierarchical interactions between both roles, which is elaborated further in this subsection.

We start by deriving the Stackelberg rule for the leader $l$. Each action in the performance map's state combinations $a_{l,p}^{q}$ is adjusted based on its potential function $\phi_L^i$ and the follower's potential function $\phi_F^i$, which uses deterministic learning techniques, as follows:
\begin{equation}\label{eq:action_gradientebased}
    a_{l,p+1}^{q} = a_{l,p}^{q}+\alpha \cdot \omega_{l}+\gamma_{l,ou},
\end{equation}
where $\omega_{l}$ represents the gradient vector for learning in~\eqref{eq:learningupdate2}:
\begin{equation}\label{eq:gradient_leaders}
    \omega_{l} = \frac{\partial \hat{\phi}_{L}}{\partial a_{l}} - \left( \frac{\partial^2 \hat{\phi}_{F}}{\partial A_{F} \partial a_{l}} \right)^T \left( \frac{\partial^2 \hat{\phi}_{F}}{\partial {A_{F}}^2} \right)^{-1} \frac{\partial \hat{\phi}_{L}}{\partial A_{F}},
\end{equation}
where $\hat{\phi}_{L}$ and $\hat{\phi}_{F}$ are the approximations of $\phi_{L}$ and $\phi_{F}$, respectively.

Next, we derive the Stackelberg rule for the follower $f$, who responds optimally to the leader's actions. The update rule is formulated as follows:
\begin{equation}\label{eq:action_grad}
    a_{f,p+1}^{q} = a_{f,p}^{q} + \alpha \cdot \frac{\partial \hat{\phi}_{f}}{\partial a_{f}} + \gamma_{f,ou}.
\end{equation}
In the above equations, $\gamma_{l,ou}$ and $\gamma_{f,ou}$ represent an Ornstein-Uhlenbeck (OU) noise term used during exploration. Since the gradient field for the follower mirrors that of the SbPG, we can utilise the gradient-based learning procedures as detailed in~\cite{Yuwono2024a}.

\subsection{Approximation of gradient descent}\label{sec:algos_2}

In practical production environments, the players in Mod-SbSG generally acquire data on the resulting potential values from their actions, but no explicit functional relationships between these values are specified, as the potential functions are inherently embedded within the system. However, to carry out the gradient updates, it is essential to have a defined functional relationship. Therefore, we approximate the potential functions, since precise potential information is required to direct the learning gradient effectively. In this study, we use polynomial regression, as proposed in~\cite{Yuwono2024c}, which is effective for continuous gradient updates. The potential function approximations for leaders and followers are given by:
\begin{equation} 
\begin{split}
    \hat{\phi}_{L}(a_{l},\mathcal{A}_{F}) = & \beta_0 + \beta_1 a_{l} + \beta_2 \mathcal{A}_{F} + \beta_3 {a_{l}}^2 + \beta_4 {\mathcal{A}_{F}}^2 + \beta_5 a_{l} \mathcal{A}_{F} + \ldots + \beta_{n+2} {a_{l}}^n \mathcal{A}_{F} ,
\end{split}
\end{equation}
\begin{equation} 
\begin{split}
    \hat{\phi}_{F}(\mathcal{A}_{L},a_{f}) = & \beta_0 + \beta_1 \mathcal{A}_{L} + \beta_2 a_{f} + \beta_3 {\mathcal{A}_{L}}^2 + \beta_4 {a_{f}}^2 + \beta_5 \mathcal{A}_{L} a_{f} + \ldots + \beta_{n+2} {\mathcal{A}_{L}}^n a_{f} ,
\end{split}
\end{equation}
where $n$ represents the degree of the polynomial regression, and $\beta = (\beta_0, \beta_1, \ldots, \beta_{n+2})$ are the coefficients computed through the ordinary least squares estimation.

\subsection{Multi-step updates for followers}\label{sec:algos_3}

In contrast to simultaneous learning, Mod-SbSG requires an alternating training methodology for leaders and followers. In simultaneous games like SbPGs, all players update their policies concurrently at each time step. Meanwhile, in Stackelberg games, according to Theorem~\ref{theorem_3}, followers are required to converge before the next update of the leader group. Hence, a multi-step update of the follower is required while leaders generally optimize their strategies in a single step. However, waiting for full convergence of followers' strategies in each training iteration can be impractical, particularly due to the extensive state spaces commonly found in manufacturing systems, which can result in excessively long training times. A feasible solution is to limit the number of gradient updates for followers per training step.

To this end, we propose to regulate the multi-step update rates for followers by introducing three different variants, such as:
\begin{enumerate}
    \item Static number of update steps
    \item Gradient magnitude thresholding for gradient-based learning
    \item Gradual reduction method for ad-hoc learning
\end{enumerate}

In the first approach, we set a parameter $\theta_g^{static}$ to specify the number of update steps for followers during each training iteration. This parameter remains fixed throughout the training process. However, a limitation of using a static number of update steps is that, as the training approaches the optimal solution, excessive exploration and updates by followers become less impactful, which leads to lengthy training times with diminishing returns.

In the second approach, we employ a dynamic number of update steps by utilizing gradient magnitude thresholding, which is particularly effective for gradient-based learning methods like A2C~\cite{Mnih2016}. We calculate the magnitude of the gradient, $||g||$, and permit followers to continue updating their policy until this magnitude falls below a predefined threshold, $\theta_g^{grad}$. Furthermore, we implement an exponential decay of the threshold $\theta_g^{grad}$ throughout the training process, governed by the decay rate $\theta_{g,decay}^{grad}$. This approach helps balance the frequency of training iterations with improved accuracy, particularly during extended training periods.

In the third approach, we implement a dynamic number of update steps by progressively decreasing update rates, which is particularly effective for gradient-based learning methods like globally interpolated gradient-based learning~\cite{Yuwono2024a} or even best response learning~\cite{Schwung2022}. We introduce a threshold, $\theta_g^{red}$, which undergoes exponential decay throughout the training period, with the rate of decay controlled by $\theta_{g,decay}^{red}$. This threshold determines the number of update steps, rounded up as necessary.

We investigate the three methods in Section~\ref{sec:res} and empirically validate our hypothesis that full convergence of followers is not required for the overall structure to converge.

\subsection{Learning mechanism}\label{sec:algos_4}

As depicted in Fig.~\ref{fig:overview}, Mod-SbSG is composed of three interconnected games. In this subsection, we explain the learning mechanism of Mod-SbSG within a dynamic system, thereby composing the derivations from the previous section.

We assume that $t$ represents the system's time step, and each player $i$ must update their action $a_{i,t}$ at each time step. However, decision-making in Mod-SbSG is role-dependent. At each time step $t$, each player $i$ first acquires the current state $s_{i,t}$ from the environment. Next, each leader $l \in i$ selects an action $a_{l,t}(s_{l,t})$ based on the current state, engaging in an SbPG among the leaders, which results in the coalition strategy $\mathcal{A}_L^t$ for the leaders. Each follower $f \in i$ then responds with an action $a_{f,t}(s_{f,t}, \mathcal{A}_L^t)$, based on both the current states and the leaders' coalition strategy. The followers also engage in SbPG, forming the coalition strategy $\mathcal{A}_F^t$. Both coalition strategies $\mathcal{A}_L^t$ and $\mathcal{A}_F^t$ are then combined through a Stackelberg game, which results in the overall set of player actions $A_t$. These actions are forwarded to the environment, which updates the state to $S_{t+1} \leftarrow S_t$ and calculates the potential values for both roles, $\phi_{L,t}$ and $\phi_{F,t}$. The process then repeats with the next time step, $t \leftarrow t+1$.

Algorithm~\ref{alg:mod_sbsg} outlines the pseudocode for Mod-SbSG. During the training of Stackelberg strategies, at each time step $t$, followers perform multi-step optimization of their policy $\pi_f$, while keeping the leaders' strategies $\mathcal{A}_L^t$ fixed, as discussed in Sec.~\ref{sec:algos_3}.

\begin{algorithm}
\caption{Basic of Mod-SbSG.}\label{alg:mod_sbsg}
\KwData{$T_{max}, \alpha, S_0, A_0$}
\For{$t=0,1,\ldots,T_{max}$}{
    \For{each leader $l$}{
        obtain $q,p$ according to $s_{l,t}$;\\
        $a_{l,p}^{q} \leftarrow \pi_l(s_{l,t})$;\\
        $a_{l,t} \leftarrow a_{l,p}^{q}$;\\
    }
    $\mathcal{A}_L^{t} = \{ a_1 \times a_2 \times \ldots \times a_L \}_t$;\\
    \For{each follower $f$}{
        obtain $q,p$ according to $s_{f,t}$;\\
        $a_{f,p}^{q} \leftarrow \pi_f(s_{f,t}, A_{L}^{t})$;\\
        $a_{f,t} \leftarrow a_{f,p}^{q}$;\\       
    }
    $\mathcal{A}_F^{t} = \{ a_1 \times a_2 \times \ldots \times a_F \}_t$;\\
    $A_t = \mathcal{A}_L^{t} \times \mathcal{A}_F^{t} $;\\
    \For{each leader $l$}{
        $\hat{\phi}_{L}(a_{l},\mathcal{A}_{F}) = \beta_0 + \beta_1 a_{l} + \beta_2 \mathcal{A}_{F} + \beta_3 {a_{l}}^2 + \beta_4 {\mathcal{A}_{F}}^2 + \beta_5 a_{l} \mathcal{A}_{F} + \ldots + \beta_{n+2} {a_{l}}^n \mathcal{A}_{F}$;\\
        $\omega_{l} = \frac{\partial \hat{\phi}_{L}}{\partial a_{l}} - \left( \frac{\partial^2 \hat{\phi}_{F}}{\partial A_{F} \partial a_{l}} \right)^T \left( \frac{\partial^2 \hat{\phi}_{F}}{\partial {A_{F}}^2} \right)^{-1} \frac{\partial \hat{\phi}_{L}}{\partial A_{F}}$;\\
        $a_{l,p+1}^{q} = a_{l,p}^{q}+\alpha \cdot \omega_{l}+\gamma_{l,ou}$;\\
    }

    \For{each follower $f$}{
        $\hat{\phi}_{F}(\mathcal{A}_{L},a_{f}) = \beta_0 + \beta_1 \mathcal{A}_{L} + \beta_2 a_{f} + \beta_3 {\mathcal{A}_{L}}^2 + \beta_4 {a_{f}}^2 + \beta_5 \mathcal{A}_{L} a_{f} + \ldots + \beta_{n+2} {\mathcal{A}_{L}}^n a_{f}$;\\
        $\omega_{f} \leftarrow \frac{\partial \hat{\phi}_{F}}{\partial a_{f}}$;\\
        $a_{f,p+1}^{q} = a_{f,p}^{q} + \alpha \cdot \frac{\partial \hat{\phi}_{f}}{\partial a_{f}} + \gamma_{f,ou}$;\\
    }    
    $S_{t+1} \leftarrow S_t$;\\
    calculate $\phi_{L,t}, \phi_{F,t}$;\\
}
\end{algorithm}

\section{Results and Discussions}\label{sec:res}

In this section, we present the results and analysis of the proposed Mod-SbSG in two different testing environments with three industrial settings. We evaluate its performance by embedding it into two different learning algorithms: (1) a globally interpolated gradient-based learning method~\cite{Yuwono2024a} and (2) the A2C algorithm~\cite{Mnih2016} from the RL domain. Additionally, we conduct an ablation study to examine the impact of varying the number of followers' update steps and the differing focuses between leaders and followers.

\subsection{Testing environments}\label{sec:res_1}

We implemented the proposed game structure of Mod-SbSG to two laboratory test belts, such as the Bulk Good Laboratory Plant (BGLP) and its larger-scale counterpart (LS-BGLP). The LS-BGLP includes a larger number of actuators and state variables, which results in a significantly higher number of players compared to the BGLP. Additionally, validation experiments were conducted on the LS-BGLP under two different industrial settings, which are sequential processes, similar to the default BGLP configuration, and serial-parallel processes.

Moreover, the BGLP and LS-BGLP environment simulation is available through the open-source frameworks MLPro\footnote{https://github.com/fhswf/MLPro}~\cite{Detlef2022a, Detlef2022b} and MLPro-MPPS\footnote{https://github.com/fhswf/MLPro-MPPS}~\cite{Yuwono2023d, Yuwono2023e}.

\subsubsection{A Bulk Good Laboratory Plant}\label{sec:bgs_1}

\begin{figure}[t]
	\centering
	\includegraphics[width=1.0\linewidth,keepaspectratio]{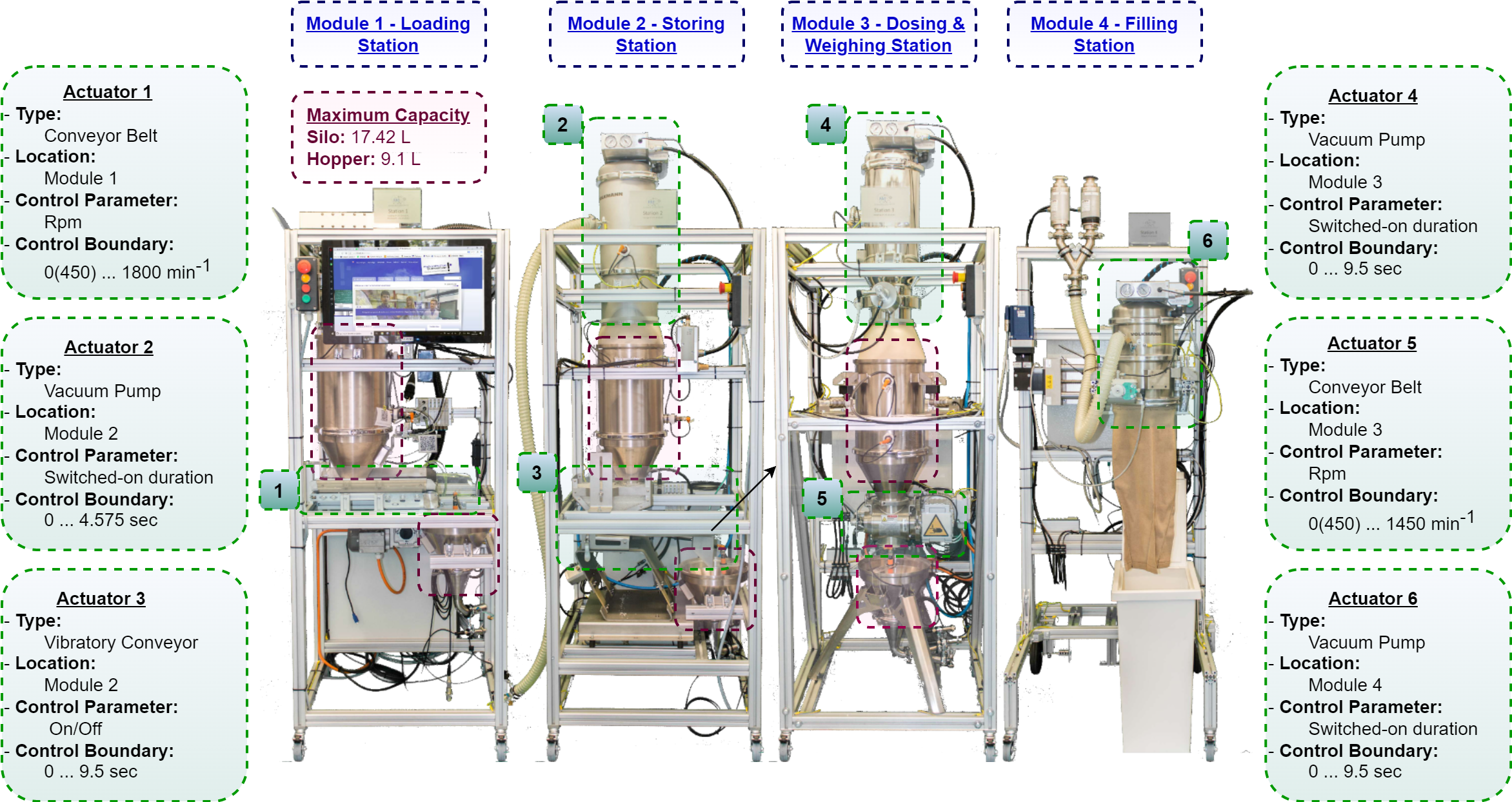}
	\caption{The Bulk Good Laboratory Plant.~\cite{Yuwono2024c}}
	\label{fig:bglp}
\end{figure}

The BGLP~\cite{Schwung2022} is a physical test belt designed to imitate a smart and adaptive manufacturing system with modular capabilities, which enables fully decentralized control. As depicted in Fig.~\ref{fig:bglp}, the primary function of the BGLP is to transport bulk goods from the initial station to the final station. The system comprises four stations, which are loading, storage, weighing, and filling. Each station features different actuators, which leads to varying control parameters across the system. Fig.~\ref{fig:bglp} provides detailed information about the actuators and reservoirs used in the BGLP. 

The primary control objective is to meet production targets while minimizing power consumption and preventing overflows or bottlenecks at any station. The contribution of each actuator to optimizing the global objective is unevenly distributed. For example, some actuators consume more power to transport the same quantity of material, while others have a higher likelihood of causing bottlenecks. This aligns well with the problem tackled in this study. Additionally, the power consumption and material transport functions exhibit non-linear behaviour with respect to their control variables.

As illustrated in Fig.~\ref{fig:bglp}, each actuator (player) $i$ is positioned between two reservoirs, typically either a silo and a hopper, or vice versa. This configuration means that each player $i$ directly impacts the two adjacent reservoirs. Hence, each leader $l$ operates within at least a two-dimensional state space, represented by the fill levels of the prior reservoir $V_i$ and the subsequent reservoir $V_{i+1}$, as follows:
\begin{align}\label{eq:states_bglp}
s_t^{l} = \{V_{l}, V_{l+1}\} \in S.
\end{align}
In contrast, for each follower $f$, this state space $s^{f}$ is expanded to include the coalition actions of the leaders, as follows:
\begin{align}\label{eq:states_bglp_followers}
S_t^{f} = \{V_{f}, V_{f+1}, A_L^t\} \in S.
\end{align}

\begin{figure}[ht]%
    \centering
    \includegraphics[width=0.65\linewidth]{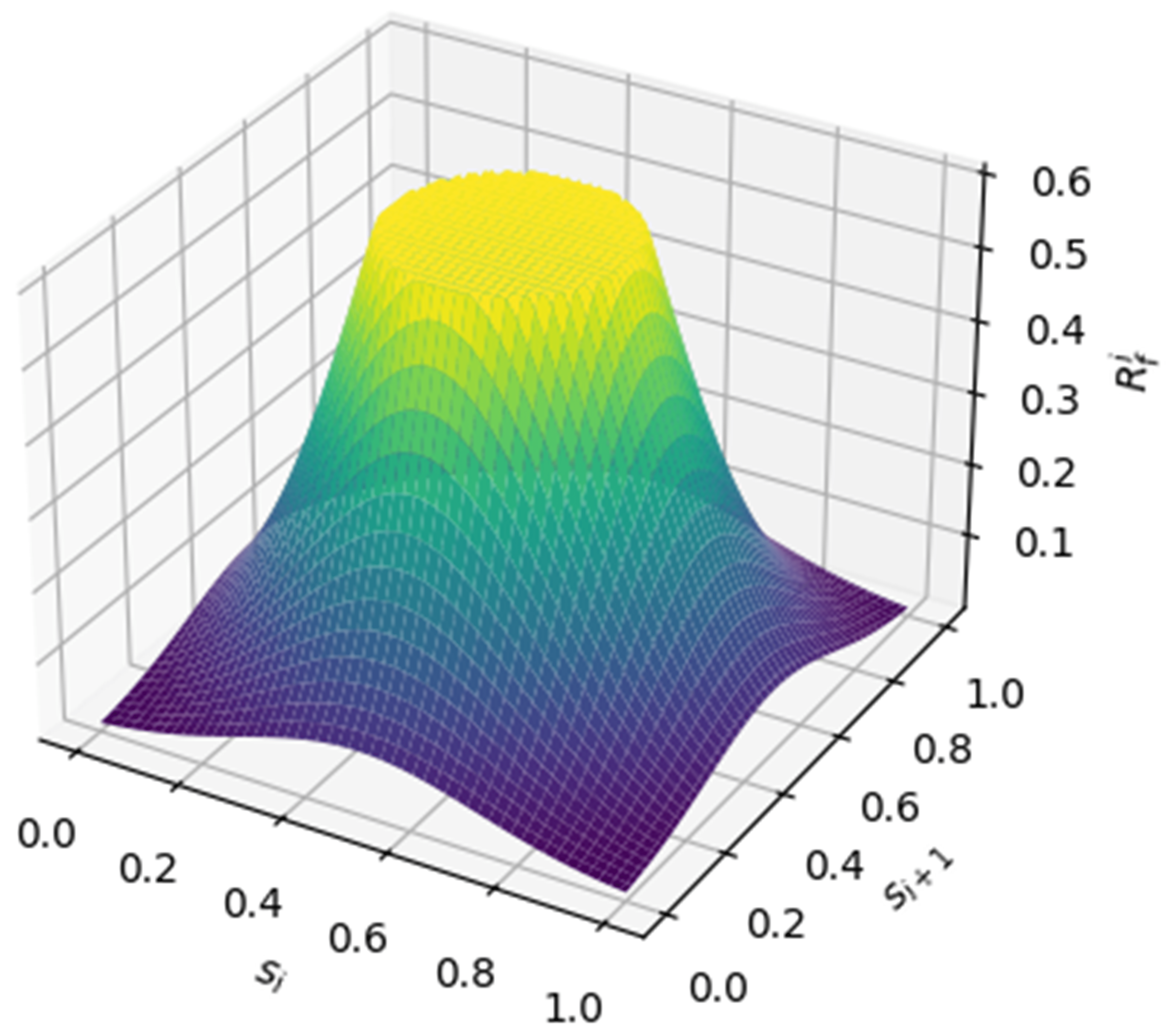}
    \caption{Output of evaluation function of the first objective using flattened bivariate normal distribution function, where $\theta_f=0.6, \sigma_p=0, \sigma_s=0, \rho=0, \mu_p=1.8, \mu_s=1.8$.}%
    \label{fig:eval}%
\end{figure}

To evaluate the performance of each player \textit{i} at time step \textit{t}, we formulate an evaluation function $E_i$, which is composed of two components, such as $E_v^i$, which handles the fill levels to prevent overflow and bottlenecks, and $E_p^i$, which addresses power consumption. The design of $E_i$ is based on a flattened version of the bivariate normal distribution function~\cite{Plackett1965}, as depicted in Fig.~\ref{fig:eval}, and formulated as follows:
\begin{equation}\label{eq:eval_volume}
    E_v^i= 
\begin{cases}
  \begin{aligned}
    \frac{1}{2\pi\sigma_p\sigma_s\sqrt{1-\rho^2}} e^{\left(-\frac{1}{2(1-\rho^2)}\left[\frac{(V_{i}-\mu_p)^2}{\sigma_p^2} - 2\rho\frac{(V_{i}-\mu_p)(V_{i+1}-\mu_s)}{\sigma_p\sigma_s} + \frac{(V_{i+1}-\mu_s)^2}{\sigma_s^2}\right]\right)},
  \end{aligned}
    & \text E_v^i \leq \theta_f
    \\
    \theta_f,              & \text{otherwise}
\end{cases}
\end{equation}
\begin{equation}\label{eq:eval_power}
    E_p^i = \frac{1}{1 + P_i},
\end{equation}
\begin{equation}\label{eq:eval_overall}
    E_i = \omega_v E_v^i + \omega_p E_p^i,
\end{equation}
where $\theta_f$ serves as a threshold to flatten the function. The parameters $\sigma_p$, $\sigma_s$, $\rho$, $\mu_p$, and $\mu_s$ originate from the bivariate normal distribution function. The weights $\omega_v$ and $\omega_p$ control the balance between the fill-level management and power consumption in the overall evaluation function. This evaluation function is applied across all experiments in this study.

\subsubsection{A Larger-Scale Bulk Good Laboratory Plant}\label{sec:bgs_2}

The LS-BGLP is an expanded version of the BGLP, which incorporates a greater number of stations, actuators, and reservoirs, along with their respective variations. Comprehensive details about the stations, actuators, and reservoirs in the LS-BGLP can be found in Tables~\ref{tab:LSBGLP_1} and ~\ref{tab:LSBGLP_2}.
\begin{table}[ht]
\renewcommand{\arraystretch}{1}
\caption{Description of the reservoirs in the LS-BGLP.}
\label{tab:LSBGLP_1}
	\centering
	\begin{tabular}{|c||c|c|c|c|}
	\hline
		\bfseries No. & \bfseries Reservoir Type & \bfseries Station & \bfseries Parameter & \bfseries Capacity\\ \hline \hline 
		1&Silo & A - Loading & Fill Level & 0...17.42L\\ \hline
		2&Hopper & A - Loading & Fill Level & 0...9.1L\\ \hline
		3&Silo & B - Feeding & Fill Level & 0...15L\\ \hline
		4&Hopper & B - Feeding & Fill Level & 0...10L\\ \hline
		5&Silo & C - Transporting & Fill Level & 0...12.5L\\ \hline
		6&Hopper & C - Transporting & Fill Level & 0...9.1L\\ \hline
		7&Mixing Silo & D - Mixing & Fill Level & 0...17.42L\\ \hline
		8&Hopper & D - Mixing & Fill Level & 0...8.0L\\ \hline
		9&Silo & E - Storing & Fill Level & 0...17.42L\\ \hline
		10&Hopper & E - Storing & Fill Level & 0...10L\\ \hline
		11&Silo & F - Weighing & Fill Level & 0...15L\\ \hline
		12&Hopper & F - Weighing & Fill Level & 0...9.1L\\ \hline
		13&Silo & G - Filling & Fill Level & 0...17.42L\\ \hline
		14&Hopper & G - Filling & Fill Level & 0...12.5L\\ \hline
		15&Big Silo & H - Batch Dosing & Fill Level & 0...30L\\ \hline
 \end{tabular}
\end{table}
\begin{table}[ht]
\renewcommand{\arraystretch}{1}
\caption{Description of the control parameters and ranges of actuators in the LS-BGLP.}
\label{tab:LSBGLP_2}
	\centering
  \begin{tabular}{|c||c|c|c|}
\hline
    \bfseries No. & \bfseries Actuator & \bfseries Parameter & \bfseries Control Range\\ \hline \hline
  1& Conveyor Belt A & Rotation Speed & $0(450)...1800rpm$\\ \hline
  2& Vacuum Pump B & On-Time Duration & 0...9.5sec\\ \hline
  3& Screw Conveyor B & Rotation Speed & $0(250)...1000rpm$\\ \hline
  4& Belt Elevator C & Rotation Speed & $0(300)...1300rpm$\\ \hline
  5& Conveyor Belt C & Rotation Speed & $0(450)...1500rpm$\\ \hline
  6& Vacuum Pump D & On-Time Duration & 0...4.575sec\\ \hline 
  7& Screw Conveyor D & Rotation Speed & $0(250)...1300rpm$\\ \hline
  8& Vacuum Pump E & On-Time Duration & 0...9.5sec\\ \hline
  9& Vibratory Conveyor E & Off / On & 0 / 1\\ \hline
  10& Belt Elevator F & Rotation Speed & $0(300)...1100rpm$\\ \hline
  11& Rotary Air Lock F & Rotation Speed & $0(450)...1450rpm$\\ \hline
  12& Bucket Elevator G & Off / On & 0 / 1\\ \hline 
  13& Dome Valve G & Open / Close & 0 / 1\\ \hline
  14& Vacuum Pump H & On-Time Duration & 0...9.5sec\\ \hline 
  \end{tabular}
\end{table}

In this study, we apply the same evaluation function as defined in Eq.~\eqref{eq:eval_overall}. We validate our proposed approaches under the LS-BGLP in two different industrial settings, as follows:
\begin{enumerate}

    \item Sequential processes: Fig.~\ref{fig:ls_bglp} depicts the LS-BGLP operating in sequential processes, where material flows sequentially through a series of stations, with each station being visited in a specific order before proceeding to the next. This setup reflects the BGLP but features a more complex arrangement. The configuration includes 14 players, each receiving two pieces of state information, which are the fill levels of the prior and subsequent reservoirs.
    \begin{figure}[ht]%
        \centering
        \includegraphics[width=1.0\linewidth]{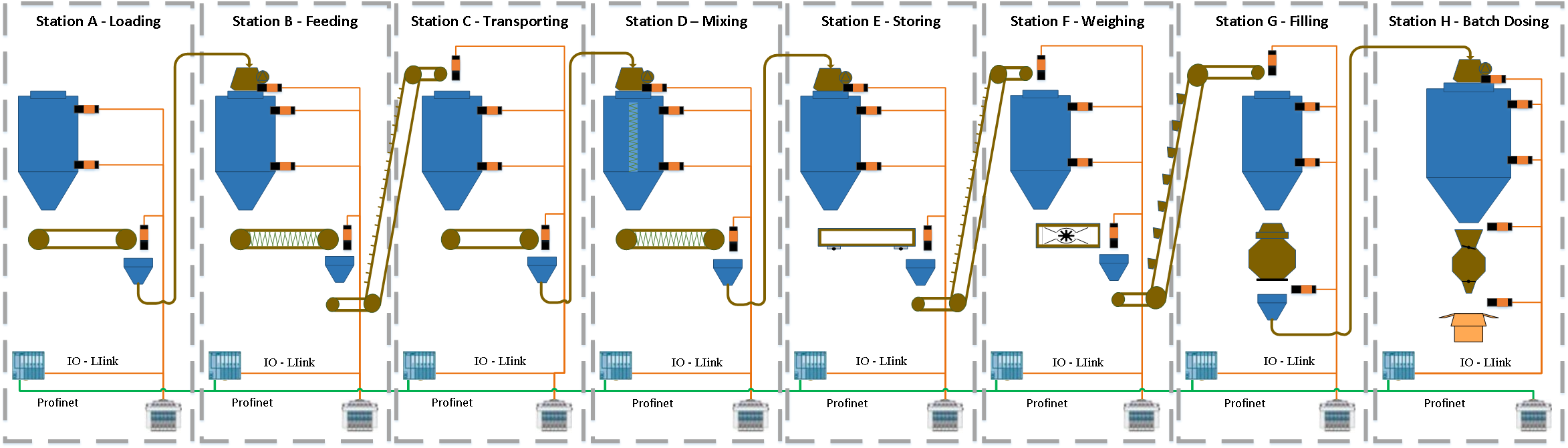}
        \caption{Larger-Scale Bulk Good Laboratory Plant~\cite{Yuwono2023c}.}%
        \label{fig:ls_bglp}%
    \end{figure}
    
    \item Serial-parallel processes: Fig.~\ref{fig:ls_bglp_sp} depicts the LS-BGLP operating in serial-parallel processes, where the sequence is modified by arranging some stations in parallel. Although the total number of players remains at 14, the state information available to some players has increased, with two reservoirs either preceding or succeeding the actuators. Additionally, certain players now interact with more than two neighbours, with one buffer being influenced by up to three actuators. This more complex configuration heightens the need for player cooperation and increases the sensitivity of action computations with respect to the global objective. Furthermore, this setup introduces a higher likelihood of experiencing overflow and bottleneck issues.
    \begin{figure}[ht]%
        \centering
        \includegraphics[width=0.8\linewidth]{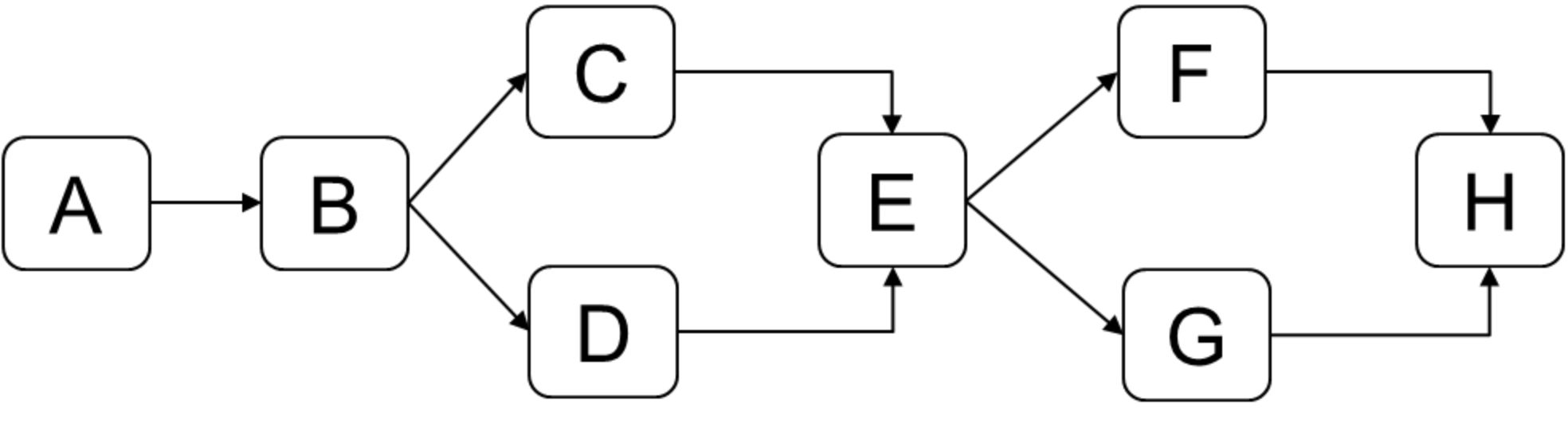}
        \caption{Modified LS-BGLP with serial-parallel processes.}%
        \label{fig:ls_bglp_sp}%
    \end{figure}
    
\end{enumerate}

\subsection{Modular State-based Stackelberg Games}\label{sec:res_2}

We set up both environments under three distinct industrial scenarios within the Mod-SbSG framework using globally interpolated gradient-based learning. Subsequently, we conducted an ablation study to another gradient-based algorithm, which is A2C from on-policy RL. The performance of each algorithm is assessed based on several key metrics, including overflow, power consumption, demand fulfilment, and evaluation values. Furthermore, we performed an ablation study to examine the differing focuses between leaders and followers.

\subsubsection{Globally interpolated gradient-based learning}\label{sec:res_2_1}

We initially configured the Mod-SbSG using globally interpolated gradient-based learning across three distinct industrial scenarios.

\paragraph{Results on the BGLP}\label{sec:res_2_1_1}\mbox{}\\
We first apply the proposed Mod-SbSG to the BGLP, using the evaluation function specified in Eq.~\eqref{eq:eval_volume}-\eqref{eq:eval_overall}. The weight parameters $\omega_v$ and $\omega_p$ in Eq.~\eqref{eq:eval_overall} are set to 1.5 and 0.1, respectively, which restricts the evaluation value $E_i$ for each player $i$ to the range of $[0,1]$. Additionally, a constant production output with a target of 0.15 L/s is maintained throughout the experiment. In this study, each cycle in Mod-SbSG corresponds to 10 seconds in the real machine. Additionally, hyperparameter tuning was performed for each experiment using Hyperopt~\cite{Bergstra2013} with a random grid search algorithm.

We then establish a baseline by designing an SbPG using gradient-based learning for the BGLP. This baseline approach involves training over 200 episodes, with each episode comprising 1,000 cycles. We validate the results by testing the algorithm in 50 episodes while maintaining the same cycle count per episode. In line with the distributed learning framework, each player operates with an individual performance map. After tuning the hyperparameters, we discretized each performance map into a 40x40 grid. Additionally, we optimized parameters such as the exploration decay rate, smoothing parameters, and other relevant settings to enhance performance.

\begin{figure}[ht]%
    \centering
    \subfloat[\centering Utilities]{{\includegraphics[width=0.975\linewidth]{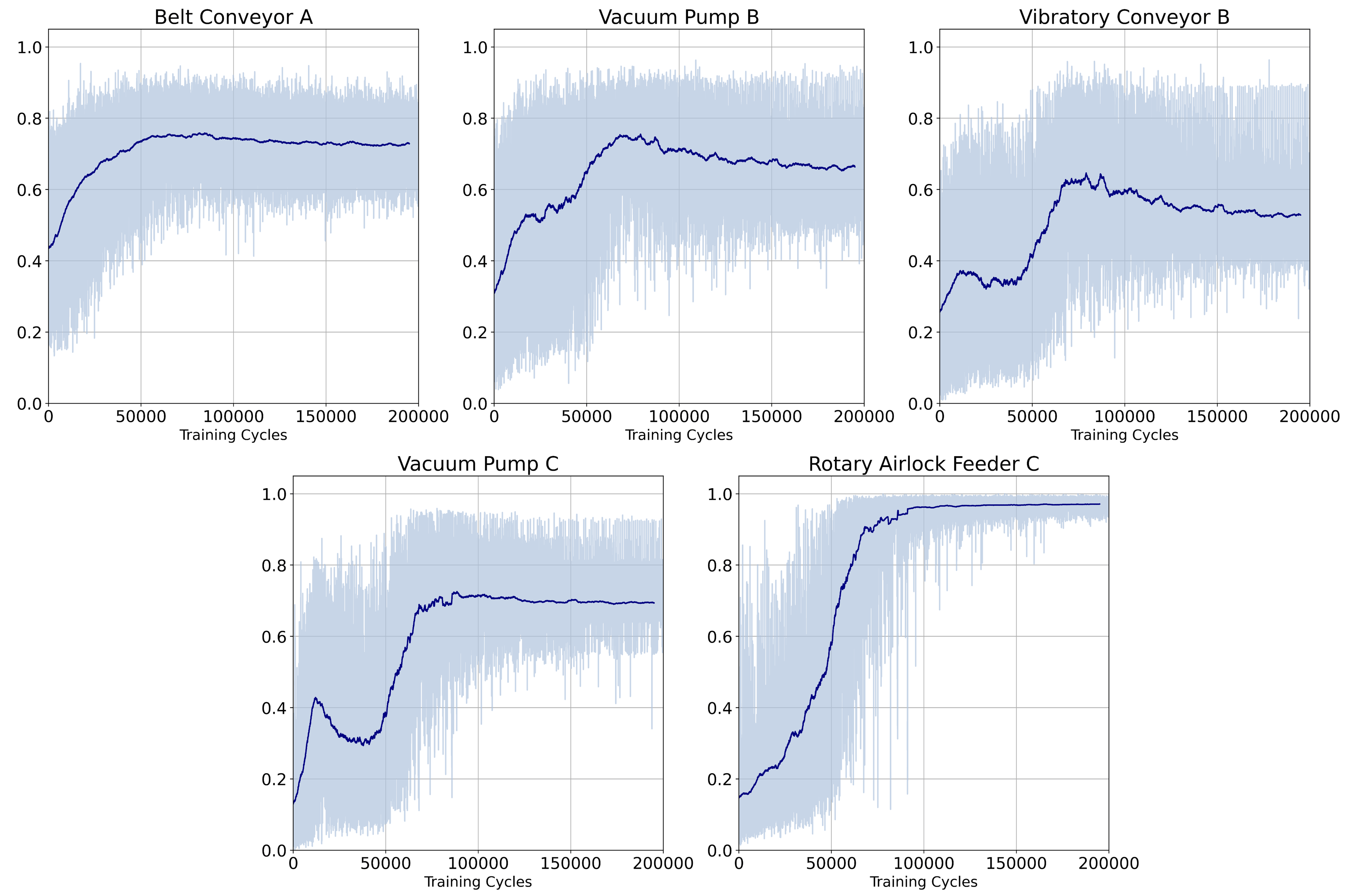} }}%
    \qquad
    \subfloat[\centering Overflow, power consumption, and production demand]{{\includegraphics[width=0.975\linewidth]{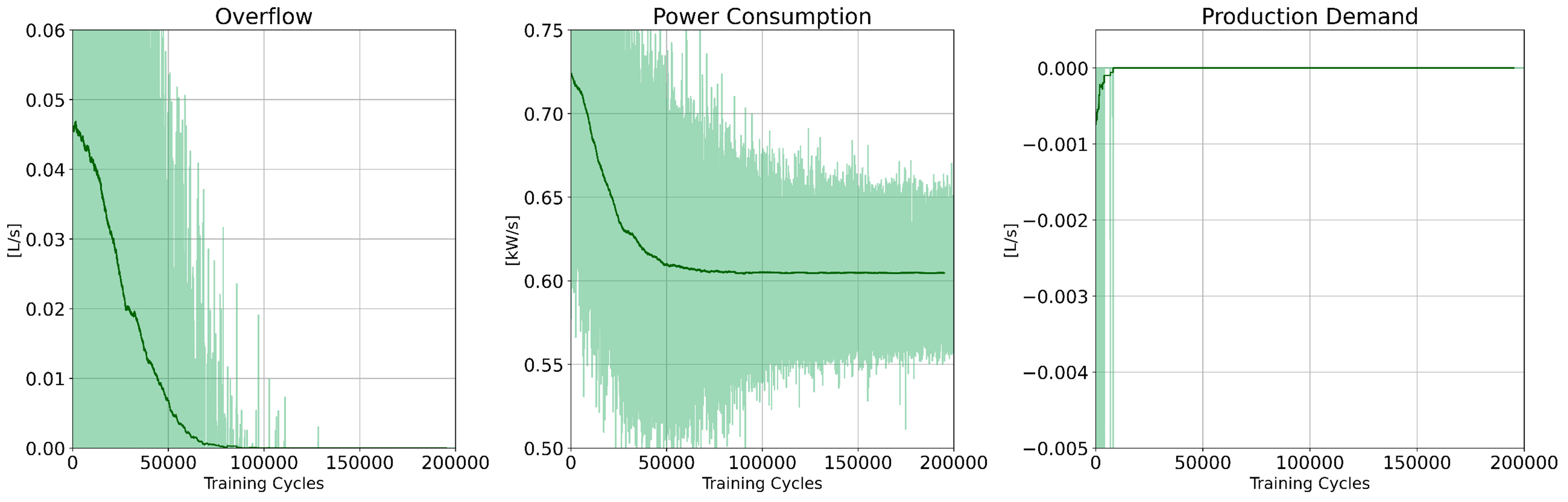} }}%
    \caption{Training results of gradient-based learning for SbPG on the BGLP.}%
    \label{fig:res_sbpg_native}%
\end{figure}

The training results for the gradient-based learning approach on SbPG for the BGLP are shown in Fig.~\ref{fig:res_sbpg_native}. During the testing phase, the production demand is consistently satisfied without overflow. The power consumption is recorded at 0.602403 kW/s, and the average potential value is 3.365761. Despite these advancements, power consumption remains relatively high, and there are occasional near-bottlenecks or overflows at certain stations.

\begin{figure}[ht]%
    \centering
    \subfloat[\centering Utilities]{{\includegraphics[width=0.975\linewidth]{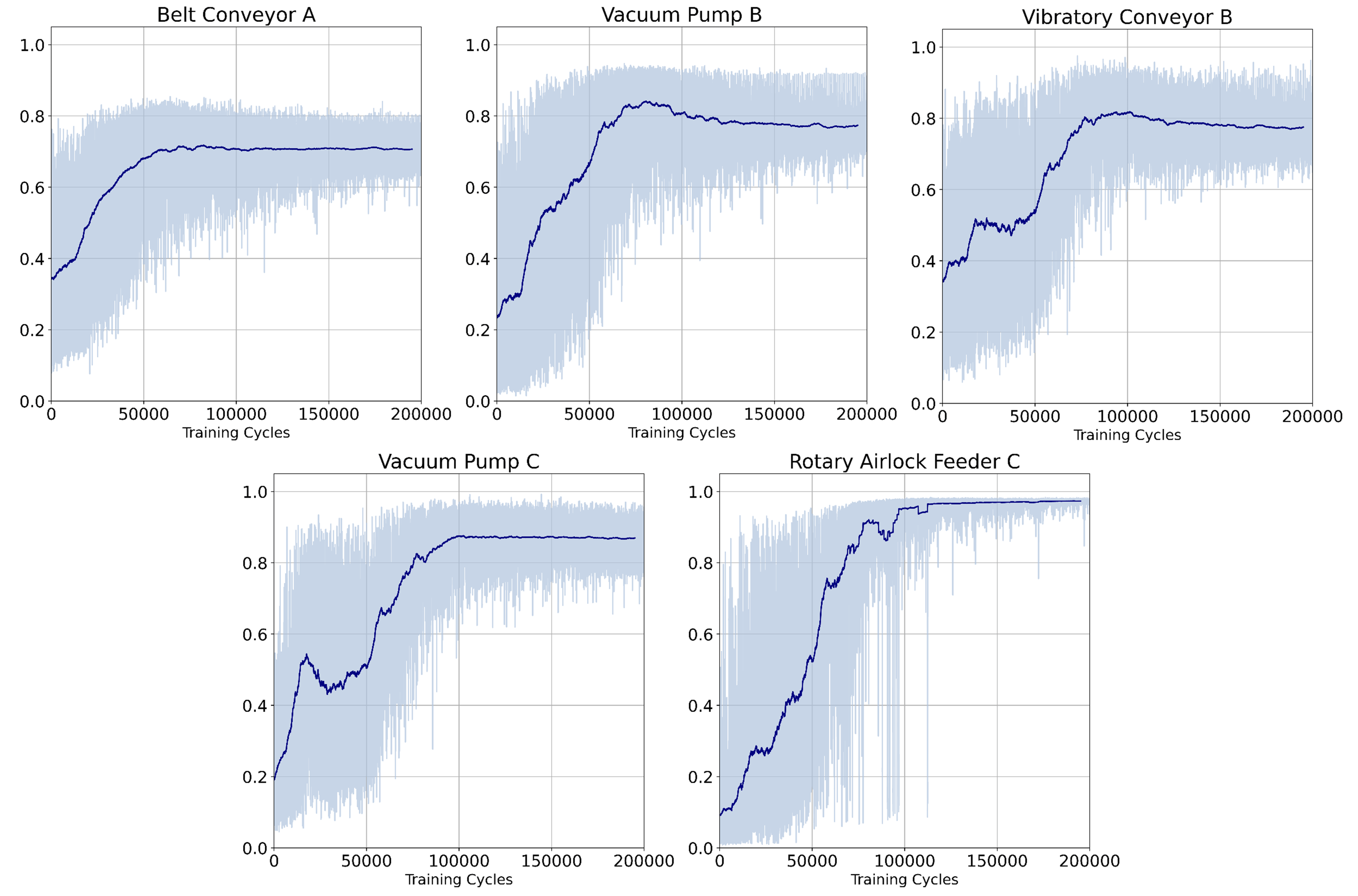} }}%
    \qquad
    \subfloat[\centering Overflow, power consumption, and production demand]{{\includegraphics[width=0.975\linewidth]{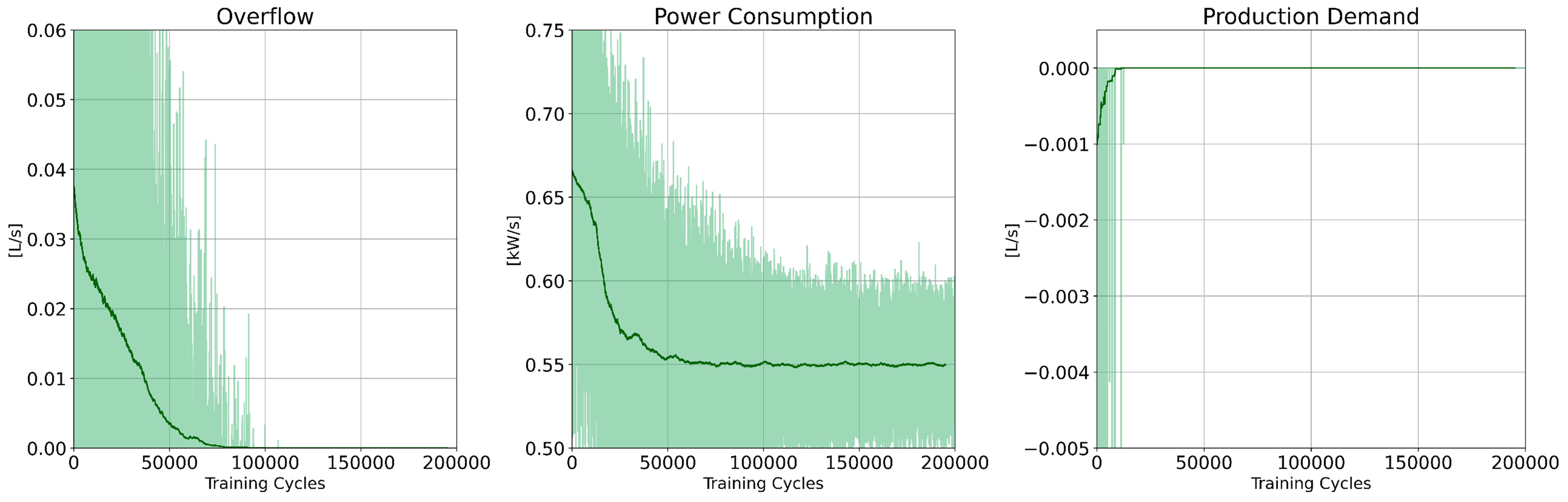} }}%
    \caption{Training results of Mod-SbSG on gradient-based learning on the BGLP, where Players 3 and 4 are the leaders.}%
    \label{fig:res_sbpg_sg}%
\end{figure}

\begin{figure}[ht]%
    \centering
    \includegraphics[width=0.50\linewidth]{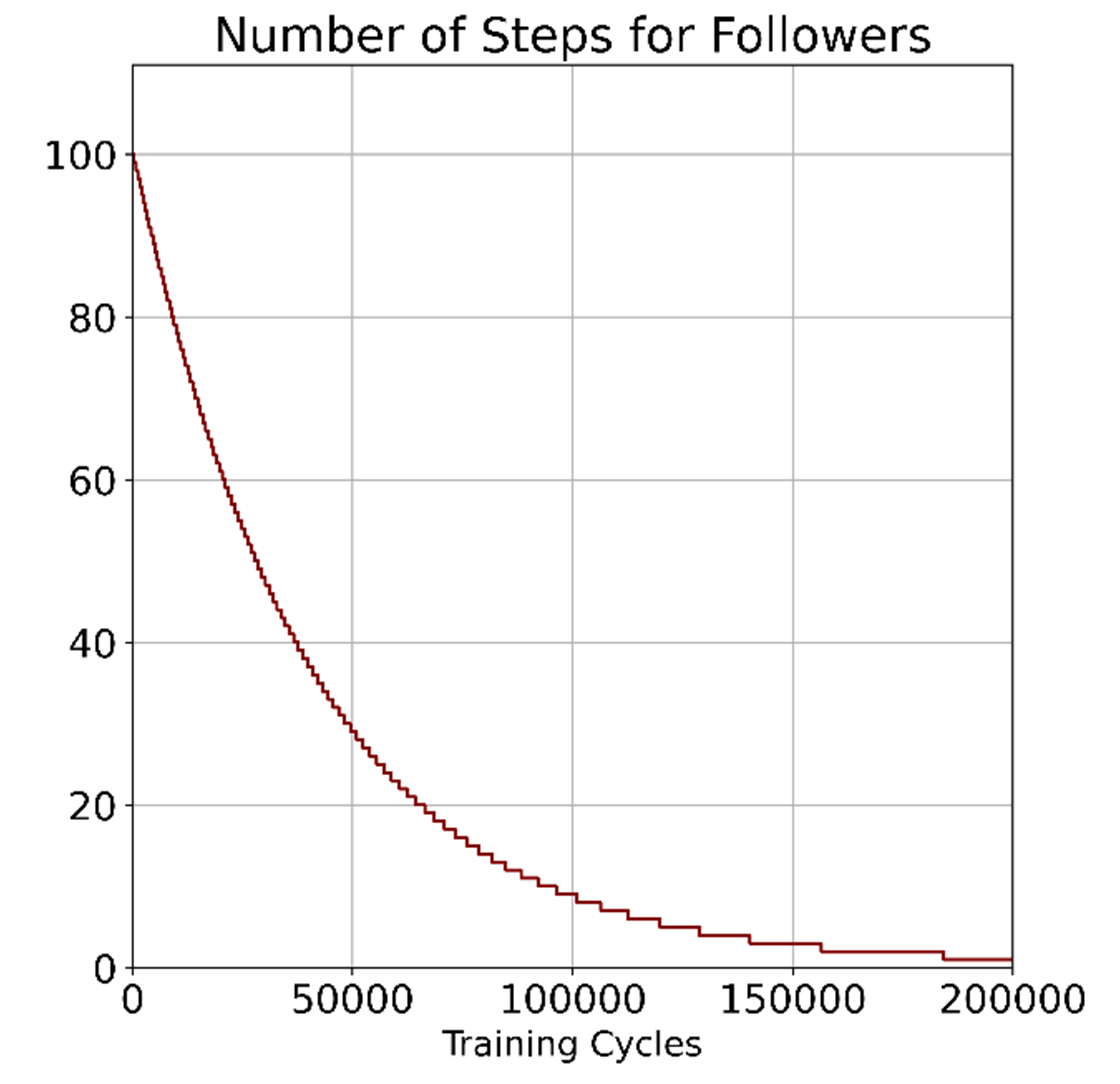}
    \caption{Number of update steps for followers using the gradual reduction method during the training of Mod-SbSG on gradient-based learning on the BGLP.}%
    \label{fig:res_sbpg_sg_2}%
\end{figure}

We afterwards implement Mod-SbSG with gradient-based learning and investigate the effect of varying the number of leaders $G$ from 1 to 3. In the Mod-SbSG framework using performance maps, followers must encode the leaders' coalition actions and stack the performance maps accordingly. For our experiments, we map these actions into a set of discrete states represented by $G\times5$, implying a minimum of 5 stacked performance maps when $G=1$. The training results with $G=2$, where players 3 and 4 act as leaders and the remaining players serve as followers, are presented in Fig.~\ref{fig:res_sbpg_sg}. Furthermore, the followers use the gradual reduction method, as shown in Fig.~\ref{fig:res_sbpg_sg_2}. This approach generates significant improvements in training outcomes compared to the native SbPG. Notably, players 2, 3, and 4 achieve higher utility values, while player 5 also maintains a high utility value. Additionally, there is a significant decrease in power consumption, no occurrence of overflow, and successful fulfilment of production demand.

We validate the proposed approach through testing episodes, and the results are summarized in Table~\ref{tab:res_modwise_sbpg_bglp}. The results also indicate that the gradual reduction method for followers is more effective than using a static number of update steps. In addition, having multiple leaders and followers ($G=2$) produces superior outcomes. Specifically, the testing results for Mod-SbSG with best response learning and $G=2$ demonstrate that production demand is satisfied, overflow is prevented, and power consumption is reduced by approximately 10.9\% compared to native SbPG. Additionally, the potential value improves by 36.5\% relative to native SbPG, which highlights significant benefits in minimizing power consumption and avoiding bottleneck situations.
\begin{table}[ht]
\renewcommand{\arraystretch}{1}
\caption{Comparisons between gradient-based learning for SbPG and Mod-SbSGs on the BGLP.}
\label{tab:res_modwise_sbpg_bglp}
	\centering
\begin{tabular}{|clccccc|}
\hline
\multicolumn{2}{|c|}{\textbf{\begin{tabular}[c]{@{}c@{}}Game\\ Structure\end{tabular}}} & \multicolumn{1}{c|}{\textbf{Leader(s)}} & \multicolumn{1}{c|}{\textbf{\begin{tabular}[c]{@{}c@{}}Demand\\ {[}L/s{]}\end{tabular}}} & \multicolumn{1}{c|}{\textbf{\begin{tabular}[c]{@{}c@{}}Power\\ {[}kW/s{]}\end{tabular}}} & \multicolumn{1}{c|}{\textbf{\begin{tabular}[c]{@{}c@{}}Overflow\\ {[}L/s{]}\end{tabular}}} & \multicolumn{1}{c|}{\textbf{Potential}} \\ \hline\hline
\multicolumn{7}{|c|}{Benchmark} \\ \hline
\multicolumn{2}{|c|}{SbPG} & \multicolumn{1}{c|}{-} & \multicolumn{1}{c|}{{\color[HTML]{680100} \textbf{0.000000}}} & \multicolumn{1}{c|}{{\color[HTML]{680100} \textbf{0.602403}}} & \multicolumn{1}{c|}{{\color[HTML]{680100} \textbf{0.000000}}} & \multicolumn{1}{c|}{{\color[HTML]{680100} \textbf{3.365761}}} \\ \hline\hline
\multicolumn{7}{|c|}{Static number of update steps  ($\theta_g^{static}=75$)} \\ \hline
\multicolumn{2}{|c|}{} & \multicolumn{1}{c|}{Pl. 4} & \multicolumn{1}{l|}{{\color[HTML]{000000} 0.000000}} & \multicolumn{1}{l|}{{\color[HTML]{000000} 0.574591}} & \multicolumn{1}{l|}{{\color[HTML]{000000} 0.000002}} & {\color[HTML]{000000} 4.196059} \\ \cline{3-7} 
\multicolumn{2}{|c|}{} & \multicolumn{1}{c|}{Pl. 3, 4} & \multicolumn{1}{l|}{{\color[HTML]{036400} \textbf{0.000000}}} & \multicolumn{1}{l|}{{\color[HTML]{036400} \textbf{0.541593}}} & \multicolumn{1}{l|}{{\color[HTML]{036400} \textbf{0.000000}}} & {\color[HTML]{036400} \textbf{4.503913}} \\ \cline{3-7} 
\multicolumn{2}{|c|}{\multirow{-3}{*}{Mod-SbSG}} & \multicolumn{1}{c|}{Pl. 2, 3, 4} & \multicolumn{1}{l|}{{\color[HTML]{000000} 0.000000}} & \multicolumn{1}{l|}{{\color[HTML]{000000} 0.575374}} & \multicolumn{1}{l|}{{\color[HTML]{000000} 0.000000}} & {\color[HTML]{000000} 3.989545} \\ \hline\hline
\multicolumn{7}{|c|}{Gradual reduction method ($\theta_g^{red}=100, \theta_{g,decay}^{red}=0.999975$)} \\ \hline
\multicolumn{2}{|c|}{} & \multicolumn{1}{c|}{Pl. 4} & \multicolumn{1}{c|}{{\color[HTML]{000000} 0.000000}} & \multicolumn{1}{l|}{{\color[HTML]{000000} 0.570559}} & \multicolumn{1}{l|}{{\color[HTML]{000000} 0.000001}} & {\color[HTML]{000000} 4.285951} \\ \cline{3-7} 
\multicolumn{2}{|c|}{} & \multicolumn{1}{c|}{Pl. 3, 4} & \multicolumn{1}{c|}{{\color[HTML]{036400} \textbf{0.000000}}} & \multicolumn{1}{l|}{{\color[HTML]{036400} \textbf{0.536387}}} & \multicolumn{1}{l|}{{\color[HTML]{036400} \textbf{0.000000}}} & {\color[HTML]{036400} \textbf{4.595327}} \\ \cline{3-7} 
\multicolumn{2}{|c|}{\multirow{-3}{*}{Mod-SbSG}} & \multicolumn{1}{c|}{Pl. 2, 3, 4} & \multicolumn{1}{c|}{0.000000} & \multicolumn{1}{l|}{0.560903} & \multicolumn{1}{l|}{0.000002} & 4.083296 \\ \hline
\end{tabular}
\end{table}

\paragraph{Results on the LS-BGLP with sequential processes}\label{sec:res_2_1_2}\mbox{}\\
Next, we apply Mod-SbSG to the LS-BGLP operating with sequential processes and compare its performance against baseline methods. In this setup, four actuators, namely Actuators 2, 3, 6, and 11, are designated as leaders. 

Table~\ref{tab:res_modwise_sbpg_lsbglp1} shows the testing results of native SbPG and Mod-SbSG on the LS-BGLP with sequential processes, in which Mod-SbSG with gradient-based learning demonstrates a significant improvement. Under this approach, production demand is consistently satisfied, and overflow is nearly eliminated, with a reduction from 0.1147357 L/s to 0.003355 L/s, representing an approximate 97.1\% decrease compared to SbPG. Additionally, power consumption is reduced by 12.4\% compared to SbPG, accompanied by an increase in potential values.
\begin{table}[ht]
\renewcommand{\arraystretch}{1}
\caption{Comparisons between gradient-based learning for SbPG and Mod-SbSG on the LS-BGLP with sequential processes.}
\label{tab:res_modwise_sbpg_lsbglp1}
	\centering
\begin{tabular}{|clccccc|}
\hline
\multicolumn{2}{|c|}{\textbf{\begin{tabular}[c]{@{}c@{}}Game\\ Structure\end{tabular}}} & \multicolumn{1}{c|}{\textbf{Leader(s)}} & \multicolumn{1}{c|}{\textbf{\begin{tabular}[c]{@{}c@{}}Demand\\ {[}L/s{]}\end{tabular}}} & \multicolumn{1}{c|}{\textbf{\begin{tabular}[c]{@{}c@{}}Power\\ {[}kW/s{]}\end{tabular}}} & \multicolumn{1}{c|}{\textbf{\begin{tabular}[c]{@{}c@{}}Overflow\\ {[}L/s{]}\end{tabular}}} & \textbf{Potential} \\ \hline\hline
\multicolumn{7}{|c|}{Benchmark} \\ \hline
\multicolumn{2}{|c|}{SbPG} & \multicolumn{1}{c|}{-} & \multicolumn{1}{c|}{{\color[HTML]{680100} \textbf{0.000000}}} & \multicolumn{1}{c|}{{\color[HTML]{680100} \textbf{1.385150}}} & \multicolumn{1}{c|}{{\color[HTML]{680100} \textbf{0.147357}}} & {\color[HTML]{680100} \textbf{21.323651}} \\ \hline\hline
\multicolumn{7}{|c|}{Gradual reduction method ($\theta_g^{red}=100, \theta_{g,decay}^{red}=0.999975$)} \\ \hline
\multicolumn{2}{|c|}{Mod-SbSG} & \multicolumn{1}{c|}{Pl. 2, 3, 6, 11} & \multicolumn{1}{c|}{{\color[HTML]{036400} \textbf{0.000000}}} & \multicolumn{1}{c|}{{\color[HTML]{036400} \textbf{1.213437}}} & \multicolumn{1}{c|}{{\color[HTML]{036400} \textbf{0.003355}}} & {\color[HTML]{036400} \textbf{25.408335}} \\ \hline
\end{tabular}
\end{table}

These improvements highlight that the Stackelberg game within Mod-SbSG significantly enhances decision-making among players in serial processes. By allowing leaders to select their actions first, and subsequently enabling followers to respond, the system achieves a more effective and coordinated approach.

\paragraph{Results on the LS-BGLP with serial-parallel processes}\label{sec:res_2_1_3}\mbox{}\\
Next, we implement Mod-SbSG to the LS-BGLP with serial-parallel processes and evaluate its performance in comparison to baseline methods, with three leaders designated as Actuators 3, 8, and 11.

Table~\ref{tab:res_modwise_sbpg_lsbglp2} compares the testing results between native SbPG and Mod-SbSG with gradient-based learning on the LS-BGLP with serial-parallel processes. While SbPG falls significantly short of meeting production demands, Mod-SbSG successfully satisfies these demands. Additionally, Mod-SbSG achieves notable reductions in overflow and power consumption by 66.6\% and 11.3\%, respectively. These improvements contribute to a higher potential value. These improvements demonstrate that Mod-SbSG substantially enhances native self-learning algorithms, especially in the context of serial-parallel processes, which are inherently more complex.
\begin{table}[ht]
\renewcommand{\arraystretch}{1}
\caption{Comparisons between gradient-based learning for SbPG and Mod-SbSG on the LS-BGLP with serial-parallel processes.}
\label{tab:res_modwise_sbpg_lsbglp2}
	\centering
\begin{tabular}{|clccccc|}
\hline
\multicolumn{2}{|c|}{\textbf{\begin{tabular}[c]{@{}c@{}}Game\\ Structure\end{tabular}}} & \multicolumn{1}{c|}{\textbf{Leader(s)}} & \multicolumn{1}{c|}{\textbf{\begin{tabular}[c]{@{}c@{}}Demand\\ {[}L/s{]}\end{tabular}}} & \multicolumn{1}{c|}{\textbf{\begin{tabular}[c]{@{}c@{}}Power\\ {[}kW/s{]}\end{tabular}}} & \multicolumn{1}{c|}{\textbf{\begin{tabular}[c]{@{}c@{}}Overflow\\ {[}L/s{]}\end{tabular}}} & \textbf{Potential} \\ \hline\hline
\multicolumn{7}{|c|}{Benchmark} \\ \hline
\multicolumn{2}{|c|}{SbPG} & \multicolumn{1}{c|}{-} & \multicolumn{1}{c|}{{\color[HTML]{680100} \textbf{-0.000012}}} & \multicolumn{1}{c|}{{\color[HTML]{680100} \textbf{1.503902}}} & \multicolumn{1}{c|}{{\color[HTML]{680100} \textbf{0.142388}}} & {\color[HTML]{680100} \textbf{19.658232}} \\ \hline\hline
\multicolumn{7}{|c|}{Gradual reduction method  ($\theta_g^{red}=100, \theta_{g,decay}^{red}=0.999975$)} \\ \hline
\multicolumn{2}{|c|}{Mod-SbSG} & \multicolumn{1}{c|}{Pl. 3, 8, 11} & \multicolumn{1}{c|}{{\color[HTML]{036400} \textbf{0.000000}}} & \multicolumn{1}{c|}{{\color[HTML]{036400} \textbf{1.333867}}} & \multicolumn{1}{c|}{{\color[HTML]{036400} \textbf{0.047587}}} & {\color[HTML]{036400} \textbf{21.814006}} \\ \hline
\end{tabular}
\end{table}

\subsubsection{Advantage Actor Critic}\label{sec:res_2_2}

We then configured the Mod-SbSG with A2C for an ablation study to investigate whether the Mod-SbSG framework can be effective when utilizing other gradient-based approaches in different domains.

\paragraph{Results on the BGLP}\label{sec:res_2_2_1}\mbox{}\\
We configure a native A2C algorithm by training it over 800 episodes, each consisting of 1,000 cycles, which is more than the episodes required for native SbPG. As with native SbPG, to validate the results, we assess the algorithm over 50 episodes, while maintaining the same number of cycles per episode, during which the policy is no longer optimized. A multilayer perceptron policy is used, and hyperparameters such as learning rate, number of steps, and gamma have been tuned. The optimal architecture for the actor and critic networks was found to be $[64,64]$ and $[32,32,16]$, respectively. Each agent is trained with a distinct actor-critic network, which reflects the distributed nature of the training process.

\begin{figure}[ht]%
    \centering
    \subfloat[\centering Rewards]{{\includegraphics[width=0.975\linewidth]{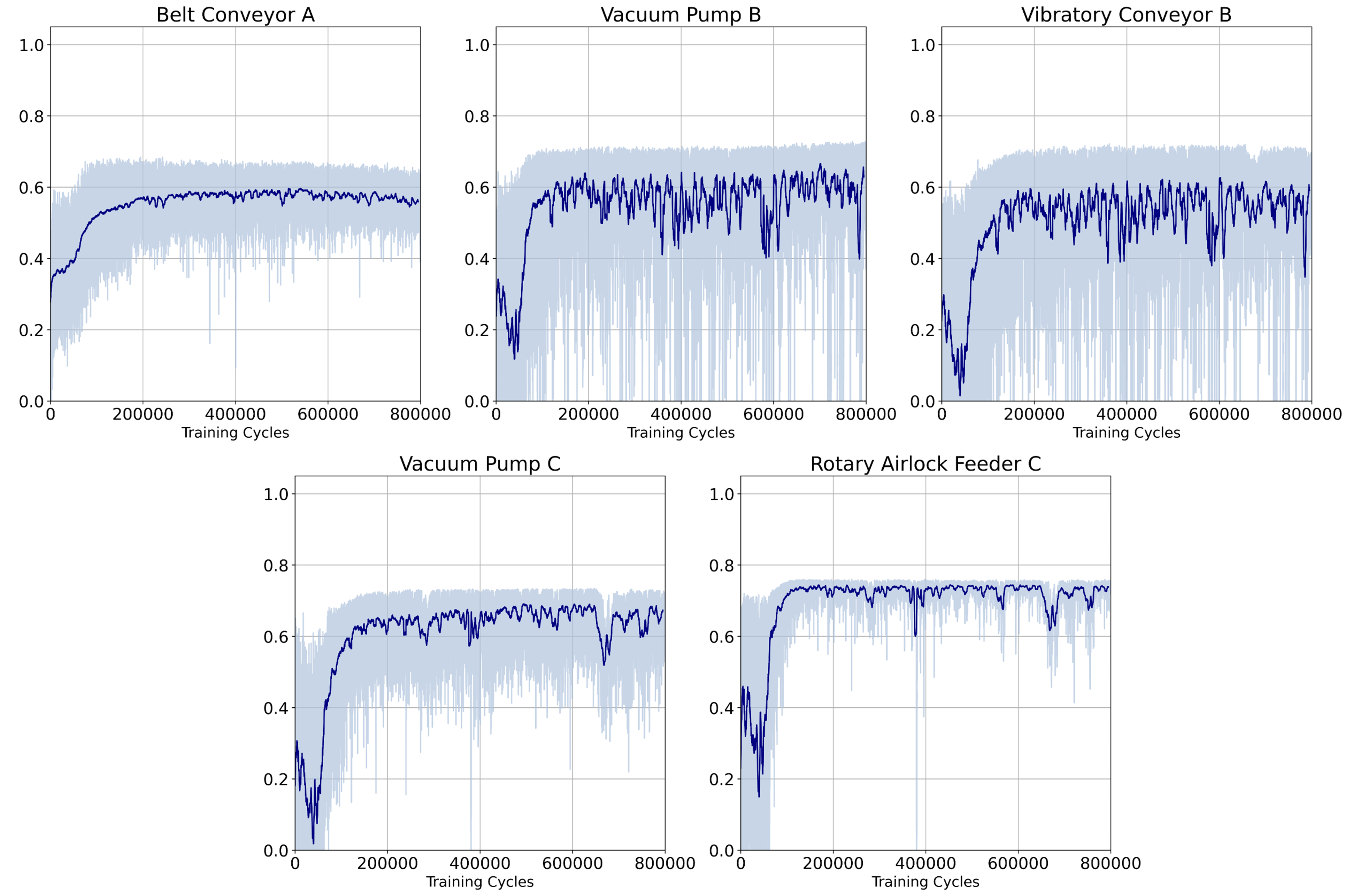} }}%
    \qquad
    \subfloat[\centering Overflow, power consumption, and production demand]{{\includegraphics[width=0.975\linewidth]{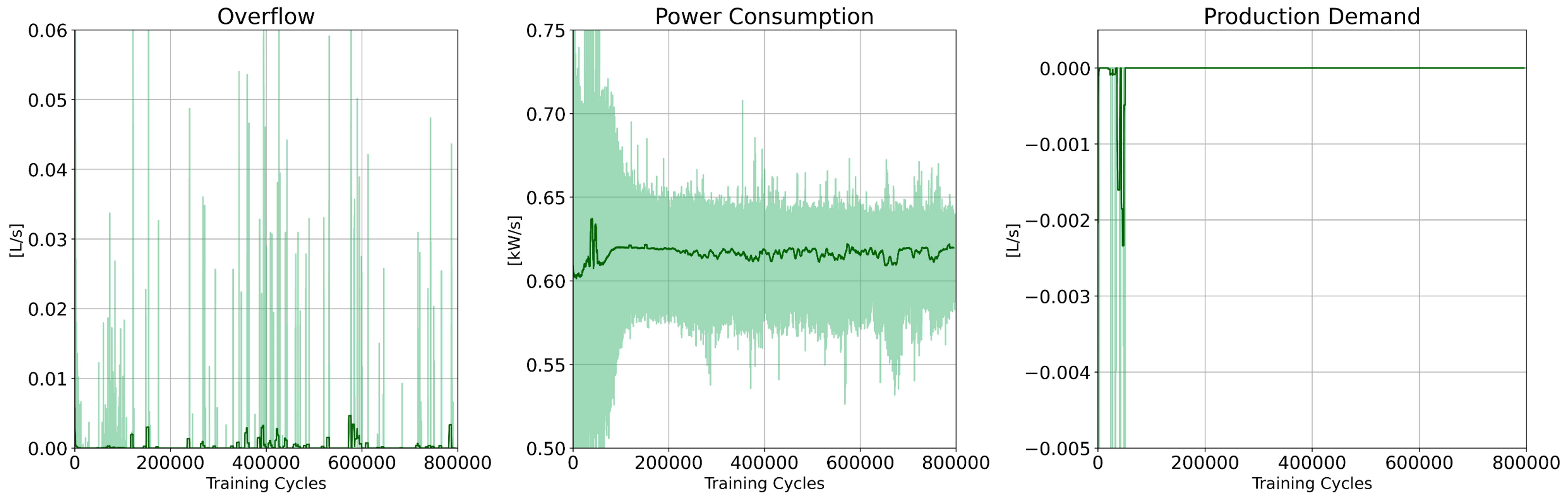} }}%
    \caption{Training results of native A2C on the BGLP.}%
    \label{fig:res_a2c_native}%
\end{figure}

Fig.~\ref{fig:res_a2c_native} describes the training outcomes of the native A2C algorithm applied to the BGLP. The graph reveals that although the agents are engaged in the learning process, they struggle to maximize their rewards effectively. The training process exhibits lesser stability and results in modestly reduced performance compared to native SbPG. During the testing phase, the system consistently meets the production demand of 0.15 L/s, yet experiences an average overflow of 0.002149 L/s and relatively high power consumption at 0.623930 kW/s. The overall reward for all agents during the testing phase averaged 2.931841. In summary, while native A2C supports the learning process for the agents, it does not achieve optimal solutions.

\begin{figure}[ht]%
    \centering
    \subfloat[\centering Rewards]{{\includegraphics[width=0.975\linewidth]{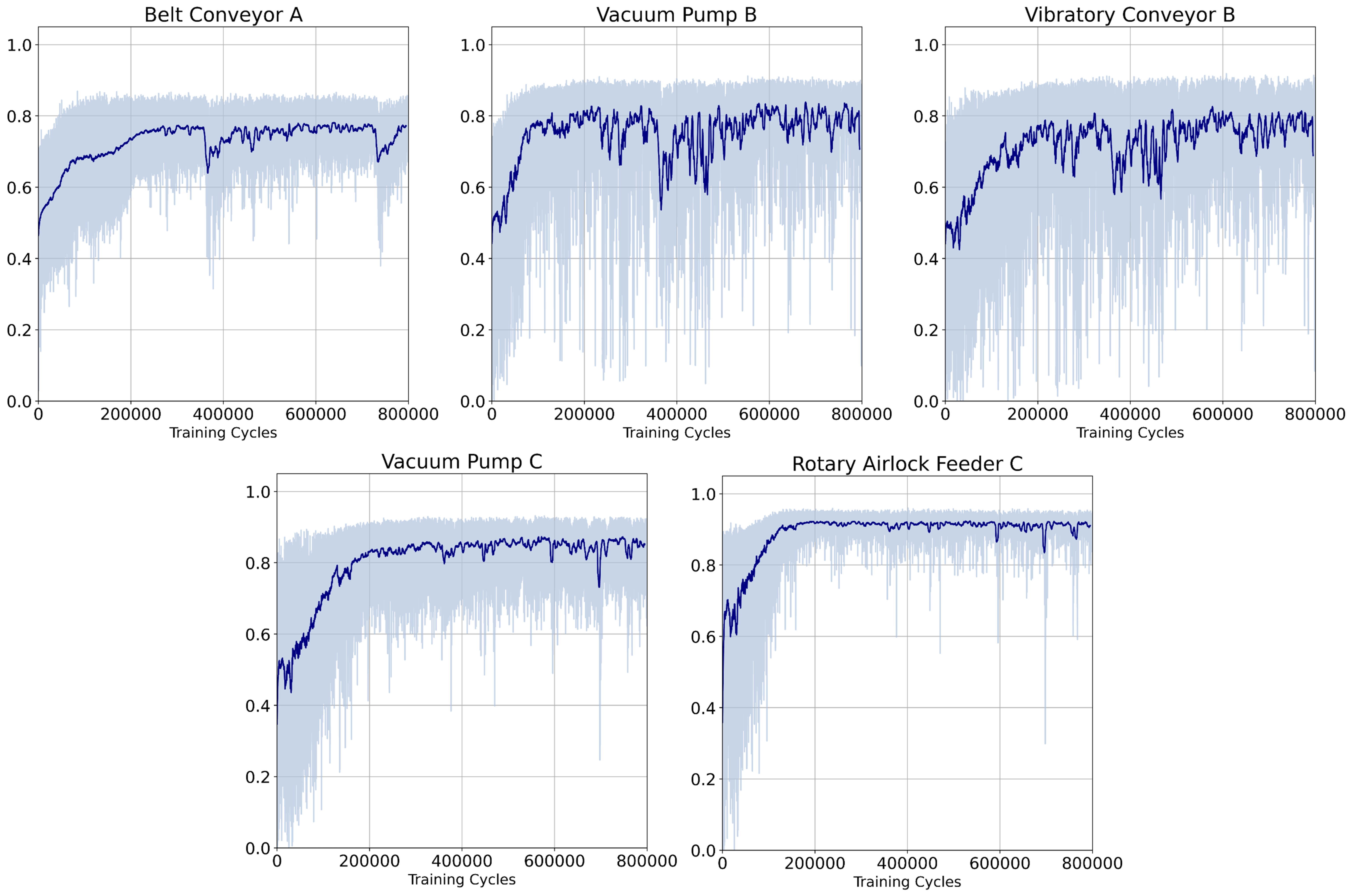} }}%
    \qquad
    \subfloat[\centering Overflow, power consumption, and production demand]{{\includegraphics[width=0.975\linewidth]{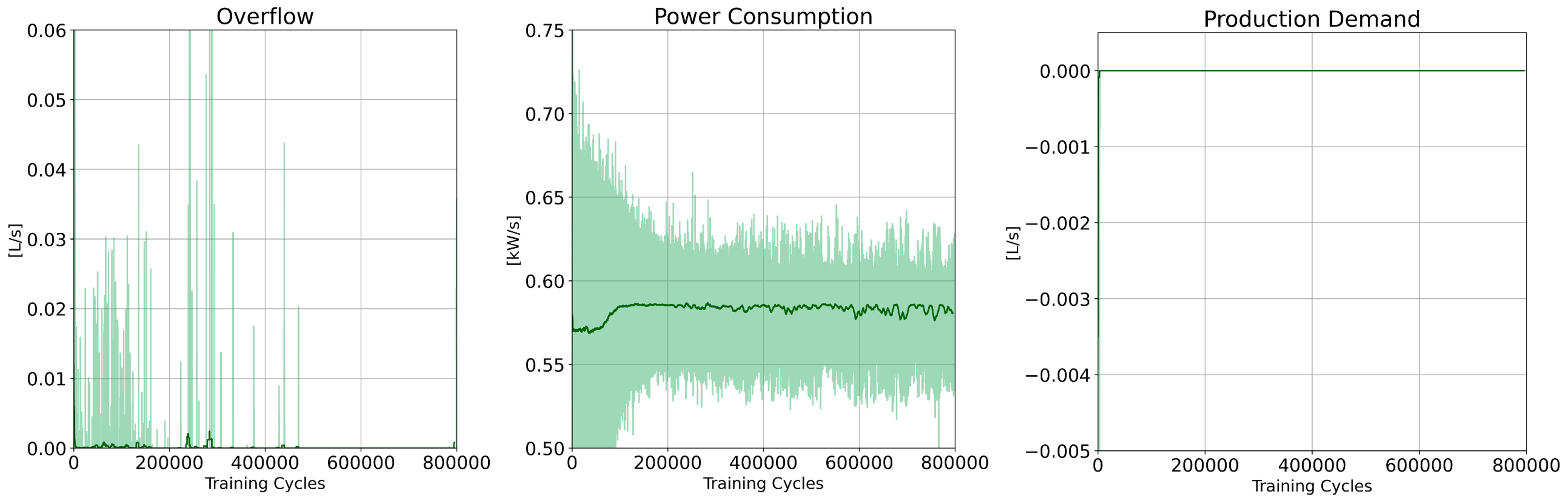} }}%
    \caption{Training results of Mod-SbSG on A2C on the BGLP, where Agent 4 is the leader.}%
    \label{fig:res_a2c_sg}%
\end{figure}

\begin{figure}[ht]%
    \centering
    \includegraphics[width=0.85\linewidth]{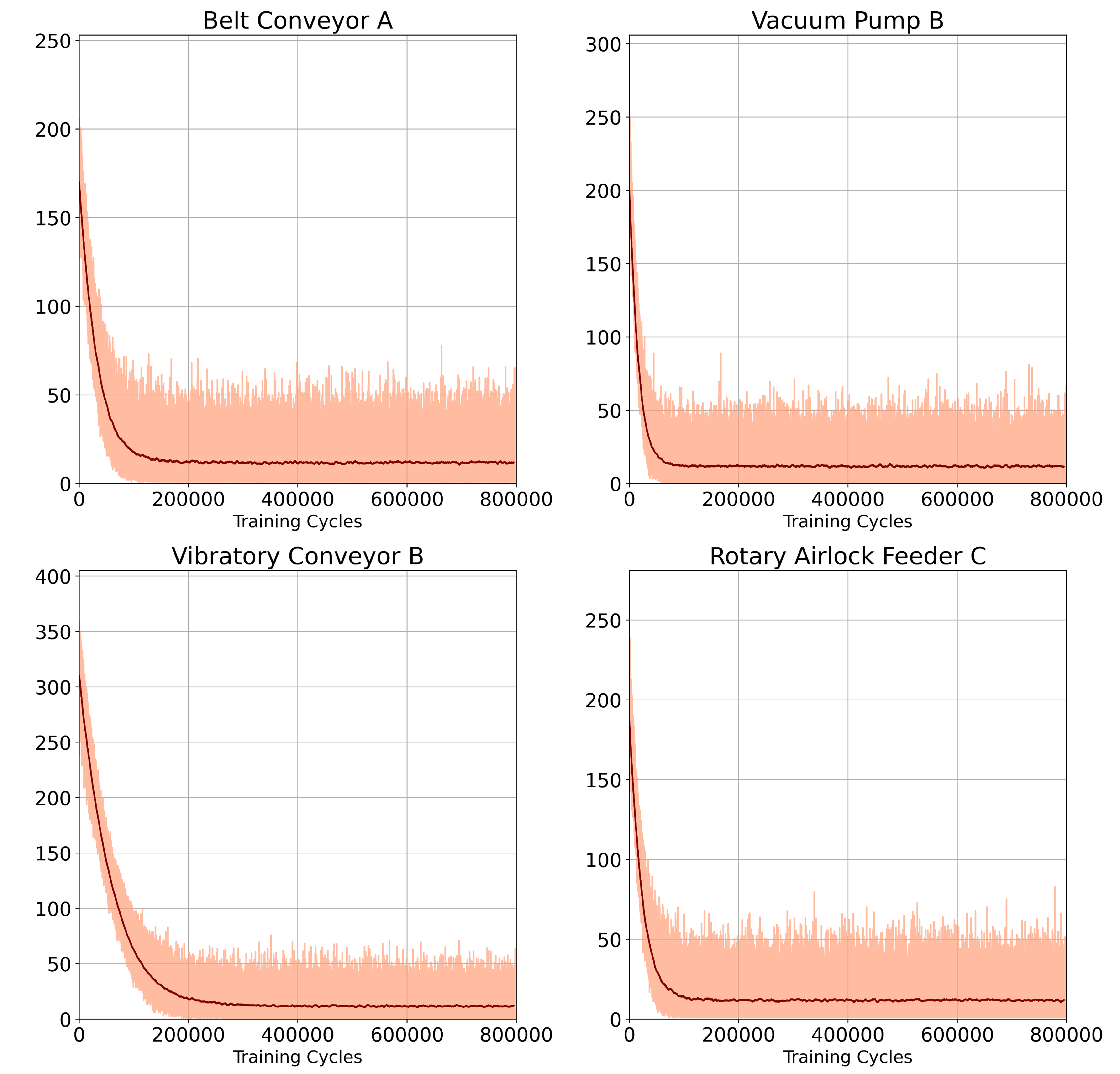}
    \caption{Number of update steps for followers using the gradient magnitude thresholding method during the training of Mod-SbSG on A2C on the BGLP.}%
    \label{fig:res_a2c_sg_2}%
\end{figure}

We then implement Mod-SbSG using A2C and test different numbers of leaders $G$ ranging from 1 to 3. The training outcomes with a single leader ($G=1$), where agent 4 takes the role of the sole leader while others act as followers are illustrated in Fig.~\ref{fig:res_a2c_sg}. The followers employ the gradient magnitude thresholding method, as shown in Fig.~\ref{fig:res_a2c_sg_2}. The graphs demonstrate an improvement in agent performance over training time. Overflow is reduced or even avoided and production demand remains fulfilled, which is a notable improvement compared to native A2C. The collaborative effort between the leader and followers contributes to a reduction in power consumption compared to native A2C. The overall performance of the BGLP controlled by Mod-SbSG on A2C is superior to native A2C, as validated in Table~\ref{tab:res_modwise_a2c_bglp}. Furthermore, it shows that the multi-step method of followers using gradient magnitude thresholding outperforms the method of using a static number of update steps. The testing results of Mod-SbSG on A2C with a single leader ($G=1$) indicate that production demand is fulfilled, overflow is avoided, and power consumption is reduced by approximately 9.65\% compared to native A2C. This improvement also leads to a higher total reward.

We then implemented Mod-SbSG using A2C and evaluated the system with varying numbers of leaders, $G$, ranging from 1 to 3. Fig.~\ref{fig:res_a2c_sg} illustrates the results for the configuration with a single leader ($G=1$), where agent 4 serves as the sole leader and the remaining agents act as followers. The followers employed the gradient magnitude thresholding method, as depicted in Fig.~\ref{fig:res_a2c_sg_2}. The graphs demonstrate noticeable improvements in agent performance over time. Specifically, overflow is minimized or eliminated, and production demand is consistently satisfied, which represents a significant advancement over native A2C.

The cooperation between the leader and followers results in reduced power consumption compared to native A2C. The performance of the BGLP controlled by Mod-SbSG with A2C surpasses that of native A2C, as detailed in Table~\ref{tab:res_modwise_a2c_bglp}. Furthermore, the multi-step optimization method for followers using gradient magnitude thresholding outperforms the static update step method. Testing results show that with a single leader ($G=1$), Mod-SbSG effectively meets production demand, avoids overflow, and reduces power consumption by approximately 9.65\% compared to native A2C. This improvement also translates into a higher total reward.
\begin{table}[ht]
\renewcommand{\arraystretch}{1.0}
\caption{Comparisons between native A2C and Mod-SbSG on A2C on the BGLP.}
\label{tab:res_modwise_a2c_bglp}
	\centering
\begin{tabular}{|clccccc|}
\hline
\multicolumn{2}{|c|}{\textbf{Algorithm}} & \multicolumn{1}{c|}{\textbf{Leader(s)}} & \multicolumn{1}{c|}{\textbf{\begin{tabular}[c]{@{}c@{}}Demand\\ {[}L/s{]}\end{tabular}}} & \multicolumn{1}{c|}{\textbf{\begin{tabular}[c]{@{}c@{}}Power\\ {[}kW/s{]}\end{tabular}}} & \multicolumn{1}{c|}{\textbf{\begin{tabular}[c]{@{}c@{}}Overflow\\ {[}L/s{]}\end{tabular}}} & \textbf{Reward} \\ \hline\hline
\multicolumn{7}{|c|}{Benchmark} \\ \hline
\multicolumn{2}{|c|}{\begin{tabular}[c]{@{}c@{}}A2C\end{tabular}} & \multicolumn{1}{c|}{-} & \multicolumn{1}{c|}{{\color[HTML]{680100} \textbf{0.000000}}} & \multicolumn{1}{c|}{{\color[HTML]{680100} \textbf{0.623930}}} & \multicolumn{1}{c|}{{\color[HTML]{680100} \textbf{0.002149}}} & {\color[HTML]{680100} \textbf{2.931841}} \\ \hline\hline
\multicolumn{7}{|c|}{Static number of update steps ($\theta_g^{static}=50$)} \\ \hline
\multicolumn{2}{|c|}{} & \multicolumn{1}{c|}{Ag. 4} & \multicolumn{1}{c|}{{\color[HTML]{036400} \textbf{0.000000}}} & \multicolumn{1}{c|}{{\color[HTML]{036400} \textbf{0.566579}}} & \multicolumn{1}{c|}{{\color[HTML]{036400} \textbf{0.000000}}} & {\color[HTML]{036400} \textbf{4.107173}} \\ \cline{3-7} 
\multicolumn{2}{|c|}{} & \multicolumn{1}{c|}{Ag. 3, 4} & \multicolumn{1}{c|}{0.000000} & \multicolumn{1}{c|}{0.589707} & \multicolumn{1}{c|}{0.000000} & 3.891637 \\ \cline{3-7} 
\multicolumn{2}{|c|}{\multirow{-3}{*}{\begin{tabular}[c]{@{}c@{}}Mod-SbSG\\ on A2C\end{tabular}}} & \multicolumn{1}{c|}{Ag. 2, 3, 4} & \multicolumn{1}{c|}{0.000000} & \multicolumn{1}{c|}{0.575975} & \multicolumn{1}{c|}{0.000000} & 3.926926 \\ \hline\hline
\multicolumn{7}{|c|}{Gradient magnitude thresholding ($\theta_g^{grad}=0.5, \theta_{g,decay}^{grad}=0.99995$)} \\ \hline
\multicolumn{2}{|c|}{} & \multicolumn{1}{c|}{Ag. 4} & \multicolumn{1}{c|}{{\color[HTML]{036400} \textbf{0.000000}}} & \multicolumn{1}{c|}{{\color[HTML]{036400} \textbf{0.563715}}} & \multicolumn{1}{c|}{{\color[HTML]{036400} \textbf{0.000000}}} & {\color[HTML]{036400} \textbf{4.123243}} \\ \cline{3-7} 
\multicolumn{2}{|c|}{} & \multicolumn{1}{c|}{Ag. 3, 4} & \multicolumn{1}{c|}{0.000000} & \multicolumn{1}{c|}{0.572834} & \multicolumn{1}{c|}{0.000000} & 3.986841 \\ \cline{3-7} 
\multicolumn{2}{|c|}{\multirow{-3}{*}{\begin{tabular}[c]{@{}c@{}}Mod-SbSG\\ on A2C\end{tabular}}} & \multicolumn{1}{c|}{Ag. 2, 3, 4} & \multicolumn{1}{c|}{0.000000} & \multicolumn{1}{c|}{0.576723} & \multicolumn{1}{c|}{0.000000} & 3.655902 \\ \hline

\end{tabular}
\end{table}

\paragraph{Results on the LS-BGLP with sequential processes}\label{sec:res_2_2_2}\mbox{}\\
We apply Mod-SbSG with A2C to the LS-BGLP operating under sequential processes. Table~\ref{tab:res_modwise_a2c_lsbglp1} shows the comparison between native A2C and Mod-SbSG with A2C for the LS-BGLP with sequential processes. Results indicate that while production demand is consistently met, overflow is significantly reduced by approximately 51.1\%, and power consumption decreases by 6.7\%. These improvements contribute to a higher overall reward.
\begin{table}[ht]
\renewcommand{\arraystretch}{1}
\caption{Comparisons between native A2C and Mod-SbSG on A2C on the LS-BGLP with sequential processes.}
\label{tab:res_modwise_a2c_lsbglp1}
	\centering
\begin{tabular}{|clccccc|}
\hline
\multicolumn{2}{|c|}{\textbf{Algorithm}} & \multicolumn{1}{c|}{\textbf{Leader(s)}} & \multicolumn{1}{c|}{\textbf{\begin{tabular}[c]{@{}c@{}}Demand\\ {[}L/s{]}\end{tabular}}} & \multicolumn{1}{c|}{\textbf{\begin{tabular}[c]{@{}c@{}}Power\\ {[}kW/s{]}\end{tabular}}} & \multicolumn{1}{c|}{\textbf{\begin{tabular}[c]{@{}c@{}}Overflow\\ {[}L/s{]}\end{tabular}}} & \textbf{Reward} \\ \hline\hline
\multicolumn{7}{|c|}{Benchmark} \\ \hline
\multicolumn{2}{|c|}{\begin{tabular}[c]{@{}c@{}}A2C\end{tabular}} & \multicolumn{1}{c|}{-} & \multicolumn{1}{c|}{{\color[HTML]{680100} \textbf{0.000000}}} & \multicolumn{1}{c|}{{\color[HTML]{680100} \textbf{1.426784}}} & \multicolumn{1}{c|}{{\color[HTML]{680100} \textbf{0.187455}}} & {\color[HTML]{680100} \textbf{19.797291}} \\ \hline\hline
\multicolumn{7}{|c|}{Gradient magnitude thresholding ($\theta_g^{grad}=0.5, \theta_{g,decay}^{grad}=0.99995$)} \\ \hline
\multicolumn{2}{|c|}{\begin{tabular}[c]{@{}c@{}}Mod-SbSG\\ on A2C\end{tabular}} & \multicolumn{1}{c|}{Ag. 2, 3, 6, 11} & \multicolumn{1}{c|}{{\color[HTML]{036400} \textbf{0.000000}}} & \multicolumn{1}{c|}{{\color[HTML]{036400} \textbf{1.330896}}} & \multicolumn{1}{c|}{{\color[HTML]{036400} \textbf{0.091731}}} & {\color[HTML]{036400} \textbf{23.170462}} \\ \hline
\end{tabular}
\end{table}

\paragraph{Results on the LS-BGLP with serial-parallel processes}\label{sec:res_2_2_3}\mbox{}\\
We introduce additional complexity to the LS-BGLP by implementing serial-parallel processes. Table~\ref{tab:res_modwise_a2c_lsbglp2} compares the performance of native A2C and Mod-SbSG with A2C in this environment. Although both approaches slightly fall short of fully meeting the production demand, Mod-SbSG with A2C demonstrates superior demand satisfaction compared to native A2C. Furthermore, Mod-SbSG with A2C achieves a substantial reduction in overflow and power consumption, with decreases of 65.5\% and 11.0\%, respectively. Additionally, the average total reward values for Mod-SbSG with A2C exceed those of native A2C by 2.950375.
\begin{table}[ht]
\renewcommand{\arraystretch}{1}
\caption{Comparisons between native A2C and Mod-SbSG on A2C on the LS-BGLP with serial-parallel processes.}
\label{tab:res_modwise_a2c_lsbglp2}
	\centering
\begin{tabular}{|clccccc|}
\hline
\multicolumn{2}{|c|}{\textbf{Algorithm}} & \multicolumn{1}{c|}{\textbf{Leader(s)}} & \multicolumn{1}{c|}{\textbf{\begin{tabular}[c]{@{}c@{}}Demand\\ {[}L/s{]}\end{tabular}}} & \multicolumn{1}{c|}{\textbf{\begin{tabular}[c]{@{}c@{}}Power\\ {[}kW/s{]}\end{tabular}}} & \multicolumn{1}{c|}{\textbf{\begin{tabular}[c]{@{}c@{}}Overflow\\ {[}L/s{]}\end{tabular}}} & \textbf{Reward} \\ \hline\hline
\multicolumn{7}{|c|}{Benchmark} \\ \hline
\multicolumn{2}{|c|}{\begin{tabular}[c]{@{}c@{}}A2C\end{tabular}} & \multicolumn{1}{c|}{-} & \multicolumn{1}{c|}{{\color[HTML]{680100} \textbf{-0.000058}}} & \multicolumn{1}{c|}{{\color[HTML]{680100} \textbf{1.644301}}} & \multicolumn{1}{c|}{{\color[HTML]{680100} \textbf{0.260679}}} & {\color[HTML]{680100} \textbf{17.860020}} \\ \hline
\multicolumn{7}{|c|}{Gradient magnitude thresholding ($\theta_g^{grad}=0.5, \theta_{g,decay}^{grad}=0.99995$)} \\ \hline\hline
\multicolumn{2}{|c|}{\begin{tabular}[c]{@{}c@{}}Mod-SbSG\\ on A2C\end{tabular}} & \multicolumn{1}{c|}{Ag. 3, 8, 11} & \multicolumn{1}{c|}{{\color[HTML]{036400} \textbf{-0.000009}}} & \multicolumn{1}{c|}{{\color[HTML]{036400} \textbf{1.463155}}} & \multicolumn{1}{c|}{{\color[HTML]{036400} \textbf{0.089865}}} & {\color[HTML]{036400} \textbf{20.810395}} \\ \hline
\end{tabular}
\end{table}

In this ablation study, we found that globally interpolated gradient-based learning generally outperforms A2C. Nevertheless, the Mod-SbSG remains effective with both gradient-based learning methods, which enhances performance in each case.

\subsection{Ablation study of focuses between leaders and followers}\label{sec:res_3}

We investigate whether the divergent priorities of leaders and followers within Mod-SbSG can improve player performance. In the BGLP, our primary goals are to avoid bottlenecks and overflow by managing fill levels and to reduce power consumption, as outlined in Eq.~\ref{eq:eval_overall}. These objectives are influenced by weight parameters $\omega_v$ and $\omega_p$. In our experimental setup, we use uniform parameters of 1.5 and 0.1 for both leaders and followers, which results in a focus of 90\% on maintaining fill levels and 10\% on reducing power consumption.

In our analysis, we explore the effects of varying priorities for leaders and followers within Mod-SbSG by using gradient-based learning with the gradual reduction method, where Players 3 and 4 serve as leaders. Table~\ref{tab:res_bot_4} provides a summary of the ablation study, showing that adjusting the focus for leaders and followers can lead to a reduction in power consumption, although the change is not highly significant. Despite this, our results suggest that differentiating the priorities of leaders and followers could be beneficial for improving multi-objective optimization outcomes.
\begin{table}[ht]
\renewcommand{\arraystretch}{1}
\caption{Ablation study of different focuses between leaders and followers in the BGLP using Mod-SbSG with gradient-based learning, where Players 3 and 4 as leaders.}
\label{tab:res_bot_4}
	\centering
\begin{tabular}{|ccccccc|}
\hline
\multicolumn{4}{|c|}{\textbf{Focuses {[}\%{]}}} & \multicolumn{1}{c|}{} & \multicolumn{1}{c|}{} &  \\ \cline{1-4}
\multicolumn{1}{|c|}{\textbf{\begin{tabular}[c]{@{}c@{}}Fill-\\ level\end{tabular}}} & \multicolumn{1}{c|}{\textbf{Power}} & \multicolumn{1}{c|}{\textbf{\begin{tabular}[c]{@{}c@{}}Fill-\\ level\end{tabular}}} & \multicolumn{1}{c|}{\textbf{Power}} & \multicolumn{1}{c|}{\multirow{-2}{*}{\textbf{\begin{tabular}[c]{@{}c@{}}Demand\\ {[}L/s{]}\end{tabular}}}} & \multicolumn{1}{c|}{\multirow{-2}{*}{\textbf{\begin{tabular}[c]{@{}c@{}}Power\\ {[}kW/s{]}\end{tabular}}}} & \multirow{-2}{*}{\textbf{\begin{tabular}[c]{@{}c@{}}Overflow\\ {[}L/s{]}\end{tabular}}} \\ \hline\hline
\multicolumn{7}{|c|}{Benchmark} \\ \hline
\multicolumn{1}{|c|}{90} & \multicolumn{1}{c|}{10} & \multicolumn{1}{c|}{90} & \multicolumn{1}{c|}{10} & \multicolumn{1}{c|}{{\color[HTML]{680100} \textbf{0.000000}}} & \multicolumn{1}{c|}{{\color[HTML]{680100} \textbf{0.536387}}} & {\color[HTML]{680100} \textbf{0.000000}} \\ \hline\hline
\multicolumn{7}{|c|}{Ablation Study} \\ \hline
\multicolumn{1}{|c|}{90} & \multicolumn{1}{c|}{10} & \multicolumn{1}{c|}{50} & \multicolumn{1}{c|}{50} & \multicolumn{1}{c|}{{\color[HTML]{036400} \textbf{0.000000}}} & \multicolumn{1}{c|}{{\color[HTML]{036400} \textbf{0.531644}}} & {\color[HTML]{036400} \textbf{0.000000}} \\ \hline
\multicolumn{1}{|c|}{70} & \multicolumn{1}{c|}{30} & \multicolumn{1}{c|}{50} & \multicolumn{1}{c|}{50} & \multicolumn{1}{c|}{-0.000598} & \multicolumn{1}{c|}{0.537707} & 0.000000 \\ \hline
\multicolumn{1}{|c|}{90} & \multicolumn{1}{c|}{10} & \multicolumn{1}{c|}{70} & \multicolumn{1}{c|}{30} & \multicolumn{1}{c|}{0.000000} & \multicolumn{1}{c|}{0.537514} & 0.000000 \\ \hline
\multicolumn{1}{|c|}{50} & \multicolumn{1}{c|}{50} & \multicolumn{1}{c|}{70} & \multicolumn{1}{c|}{30} & \multicolumn{1}{c|}{-0.001276} & \multicolumn{1}{c|}{0.541782} & 0.000532 \\ \hline
\multicolumn{1}{|c|}{70} & \multicolumn{1}{c|}{30} & \multicolumn{1}{c|}{90} & \multicolumn{1}{c|}{10} & \multicolumn{1}{c|}{0.000000} & \multicolumn{1}{c|}{0.547912} & 0.000004 \\ \hline
\multicolumn{1}{|c|}{50} & \multicolumn{1}{c|}{50} & \multicolumn{1}{c|}{90} & \multicolumn{1}{c|}{10} & \multicolumn{1}{c|}{-0.000009} & \multicolumn{1}{c|}{0.540603} & 0.000000 \\ \hline
\end{tabular}
\end{table}

\section{Conclusions}\label{sec:conclusion}

We introduce a novel game structure, Mod-SbSG, designed to facilitate leader-follower configurations in a distributed manner, and adaptable to various self-learning algorithms. This structure emphasizes self-optimization in multi-agent modular manufacturing systems and comprises three different games, including an SbPG among leaders, an SbPG among followers, and a Stackelberg game for leader-follower interactions. The effectiveness of Mod-SbSG is validated across three different industrial settings, which are the BGLP, the LS-BGLP with sequential processes, and the LS-BGLP with serial-parallel processes. In these experiments, Mod-SbSG consistently improves learning algorithm performance, which reduces overflow by up to 97.1\% compared to baseline methods and achieves a notable 5-13\% reduction in power consumption. These improvements are reflected in significant increases in potential values. Additionally, we explore various configurations of Mod-SbSG, including scenarios with single or multiple leaders, different prioritization for leaders and followers, and the regulation of follower update rates throughout the training process.

Our future work will focus on advancing Mod-SbSG to tackle constrained optimization problems. We also plan to enhance the gradient-based learning component by integrating auto-concentric performance maps. Additionally, we aim to apply Mod-SbSG to a wider range of self-learning domains, including evolutionary algorithms and model-based learning.

\section*{Declaration of Competing Interest}
The authors declare that they have no known competing financial interests or personal relationships that could have appeared to influence the work reported in this paper.

\section*{Acknowledgements}
We would like to express our gratitude to our colleagues from the Department of Automation Technology and Learning Systems at South Westphalia University of Applied Sciences for their valuable feedback and insights during this research. Additionally, we extend our thanks to all the contributors of MLPro\footnote{https://github.com/fhswf/MLPro/graphs/contributors} and MLPro-MPPS\footnote{https://github.com/fhswf/MLPro-MPPS/graphs/contributors}, for providing us with their open-source machine learning framework.

\bibliographystyle{elsarticle-num}
\bibliography{sample}

\end{document}